\documentclass[review]{elsarticle}
\usepackage{booktabs} 
\usepackage{graphicx}
\usepackage{amsfonts}
\usepackage{dsfont}
\usepackage{graphicx, epsfig,amsmath,amssymb,latexsym,graphics,float}
\usepackage{setspace,subfig}
\usepackage{citesort}
\usepackage{float}
\usepackage[ruled,vlined,linesnumbered]{algorithm2e}
\usepackage{booktabs}
\usepackage{lipsum}
\usepackage{multicol}
\usepackage{url}
\usepackage{multirow}
\usepackage[utf8]{inputenc}
\usepackage[small]{caption}
\usepackage{graphicx}
\usepackage{amsmath}
\usepackage{booktabs}
\usepackage{algorithmic}
\begin{document}
\begin{frontmatter}
\title{Sequential online prediction in the presence of outliers and change points: an instant temporal structure learning approach
}
\author{$^{\star}$Bin Liu$^{1,2}$, Yu Qi$^{3}$ and Ke-Jia Chen$^{1,2}$}
\address{$^{1}$ School of Computer Science, Nanjing University of Posts and Telecommunications\\
$^{2}$ Jiangsu Key Lab of Big Data Security $\&$ Intelligent
Processing\\
$^{3}$College of Computer Science and Technology, Zhejiang University\\
$^{\star}$Corresponding author. Email: bins@ieee.org}
\begin{abstract}
In this paper, we consider sequential online prediction (SOP) for streaming data in the presence of outliers and change points. We propose an INstant TEmporal structure Learning (INTEL) algorithm to address this problem.
Our INTEL algorithm is developed based on a full consideration of the duality between online prediction and anomaly detection. We first employ a mixture of weighted Gaussian process models (WGPs) to cover the expected possible temporal structures of the data. Then, based on the rich modeling capacity of this WGP mixture, we develop an efficient technique to instantly learn (capture) the temporal structure of the data that follows a regime shift. This instant learning is achieved only by adjusting one hyper-parameter value of the mixture model. A weighted generalization of the product of experts (POE) model is used for fusing predictions yielded from multiple GP models. An outlier is declared once a real observation seriously deviates from the fused prediction. If a certain number of outliers are consecutively declared, then a change point is declared. Extensive experiments are performed using a diverse of real datasets. Results show that the proposed algorithm is significantly better than benchmark methods for SOP in the presence of outliers and change points.
\end{abstract}

\end{frontmatter}
\section{Introduction}
Driven by the fast development of communication, sensor and storage technologies, streaming data abounds in many application areas such as transportation \cite{biem2010ibm}, manufacturing \cite{harvey1990forecasting}, network security \cite{condon2008analysis}, agriculture monitoring \cite{eerens2014image} and medical diagnosis \cite{rasmussen2011inference}. In this paper, we are concerned with the issue of sequential online prediction (SOP) for streaming data. An online real-time prediction algorithm is critical for many use cases, e.g., preventative maintenance, fraud prevention, and real-time monitoring. Compared with batch processing methods that need to access each data point repeatedly, an online prediction algorithm access each data point only once, and thus is beneficial for saving costs in computation and storage.

In many time-series data, outliers and change points exist. An outlier is typically a single observation whose statistical property is independent of and different from that of the rest data. Different from outliers, a change point stands for a data point in the time-series, at which an endogenous regime shift of the system happens. That says, after a change point, the statistics of the system changes. It is an issue also known as concept drift.

If not appropriately dealt with, the presence of outliers and change points can lead to detrimental effects on the prediction. To clarify the necessity and importance for outlier and change point detection, let consider a sensor network based real-time monitoring scenario. Sensor failures shall produce outliers \cite{liu2015toward,liu2017state}, while, when an abnormal event like a forest fire, a leakage of poisonous gas or debris flow happens, it will lead to a regime shift, corresponding to a change point in sensor measurements \cite{wang2017online}. In this scenario, undiscovered outliers shall produce unreliable predictions, and, more crucially, if a regime shift is not detected and addressed well in time, it can lead to a disaster.

Approaches to outlier and change point detection can be roughly categorized into two classes, namely retrospective methods and online approaches. For the former class, the full dataset is available for analysis, and the locations of anomalies are identified in a batch mode \cite{xuan2007modeling,barry1993bayesian}. Many previous approaches to anomaly detection are retrospective, while the batch processing feature makes them not suitable for use in an online prediction system.

Compared with retrospective methods, online approaches have received less attention. This is possibly due to that the online setting is quite different from traditional situations that make traditional approaches inapplicable.
Bayesian modeling and inference approaches have been explored for developing online change point detection methods in e.g., \cite{adams2007bayesian,turner2009adaptive,fearnhead2007line,ray2002bayesian,chandola2011gaussian}.
In the Bayesian online change point detection (BOCPD) algorithm of \cite{adams2007bayesian,turner2009adaptive}, authors design an underlying predictive model (UPM) and a hazard function to describe uncertain factors regarding the \emph{run length}, namely the time since the last change point. For BOCPD, it assumes that all data points within a regime are identically and independently distributed (iid) according to a specific distribution, e.g., Gaussian. This assumption makes BOCPD a pure change point detection method without the capability to do online prediction. In \cite{saatcci2010gaussian}, the Gaussian process (GP) is introduced into the BOCPD framework to exploit the temporal structure of the data, while its central aim is still to improve the change point detection performance.

Outliers and change points are usually separately considered in the literature (with few exceptions in e.g., \cite{yamanishi2002unifying}), under the name of outlier detection, fault detection, and change point detection. In this paper, we consider them together and propose an SOP algorithm that is robust to both outliers and change points. Our algorithm is based on the Gaussian process time-series (GPTS) model. GPs are Bayesian nonparametric models that have widely used for approximating complex nonlinear functions. It has been proved that the prediction performance of GP is comparable to that of artificial neural networks (ANN) \cite{neal2012bayesian}. Further, GP has two merits compared with ANN. First, the probabilistic nature of GP can give us a byproduct, i.e., an uncertainty measure for the prediction it makes. This is a desirable property for human operators to make decisions. Second, as a generative model, GP is more interpretable.

GPTS models have recently been studied for developing robust SOP methods \cite{roberts2013gaussian,osborne2012real,vanhatalo2009gaussian,osborne2011machine,garnett2010sequential}.
For example, in \cite{vanhatalo2009gaussian}, the authors take into account the presence of faulty observations in the GPTS model by using a heavy-tailed student's $t$ likelihood function. A similar idea has been used in \cite{osborne2011machine}, which takes account of faulty data by a Gaussian distribution with a very wide variance. In \cite{garnett2010sequential}, the authors design a non-stationary kernel function to take account of the appearance of change points. Although these methods are powerful, when using them, one has to pay some price, that is the significantly increased complexity in the inference. This is due to the lack of an analytically tractable inference algorithm for those models. Specifically, for the student's $t$ model of \cite{vanhatalo2009gaussian}, approximate methods such as Gibbs sampling \cite{geweke1993bayesian} and variational methods \cite{tipping2005variational} are needed for inference. For the model of \cite{garnett2010sequential}, a complex Bayesian quadrature procedure is required for calculating the predictive distribution.

In this paper, we propose an INstant TEmporal structure Learning (INTEL) algorithm for SOP, which has desirable features as follows:
\begin{enumerate}
\item Different from the aforementioned GPTS based methods, our INTEL algorithm allows closed-form inference and prediction, and thus is more computationally efficient and easier to code;
\item The INTEL algorithm is robust to both outliers and change points;
\item As a GP based algorithm, INTEL inherits all merits of GPs. For example, it can produce an uncertainty measure for each prediction it makes. It also has the desirable interpretability;
\item As an SOP algorithm, it can also provide real-time anomaly detections as a byproduct.
\end{enumerate}
To the best of our knowledge, INTEL is the only algorithm in the literature that owns all the above features.
The other contributions of this paper are summarized as follows:
\begin{enumerate}
\item We present a mixture modeling approach to pre-cover temporal structures of unobserved time-series data based on a template model trained with a relatively small number of observed data points. Using this method, we obtain a mixture model that has a much richer modeling capacity than the template model.
\item Based on the rich modeling capacity of the aforementioned mixture model, we propose an efficient approach to quickly capture the temporal structure of the new regime upon a change point is detected. We term this mechanism of fast temporal structure capturing as instant learning. Striking different from traditional machine learning (ML) methods, instant learning emphasizes the use of prior knowledge and does not require any training dataset.
\item We present a weighted generalization of the POE model of \cite{hinton2002training} for fusing predictions yielded from multiple GPTS models.
\item We use a bunch of real datasets to evaluate the performance of our method. Results demonstrate the superiority of our method.
\end{enumerate}

The remainder of this paper is organized as follows. In Section \ref{sec:gp}, we present the GPTS model we use for developing the proposed INTEL algorithm. In Section \ref{sec:INTEL}, we introduce INTEL in detail and provide a formal analysis of its computational complexity. In Section \ref{sec:connection}, we discuss its connections to relevant works in the literature. In Section \ref{sec:experiment}, we evaluate the performance of INTEL using a bunch of real datasets. Finally, we conclude the paper in Section \ref{sec:con}.
\section{Sequential online prediction with GP}\label{sec:gp}
In this section, we briefly introduce the GPTS model used here. The aim is to fix notations and introduce the necessary background information for presenting the INTEL algorithm later in Section \ref{sec:INTEL}. For more details on GP and its applications in time-series prediction, readers are referred to \cite{roberts2013gaussian,williams2006gaussian}.
\subsection{GP}
Let start by introducing the GP. GP is a probability distribution defined on a function. Consider a function $f$ drawn from a GP as follows
\begin{equation}\label{eqn:gp_def}
y=f(x), f \sim \mathcal{GP}\left(\mu, k_{\theta}\right),
\end{equation}
where $x$ is the input of the function, $y$ is the output, $\mathcal{GP}\left(\mu, k_{\theta}\right)$ denotes a GP with mean function $\mu(\cdot)$ and covariance kernel function $k_{\theta}(\cdot,\cdot)$. Here $\theta$ denotes the hyper-parameter of the kernel function. Given any two (arbitrary) input locations, say $x_i$ and $x_j$, the kernel function defines the covariance element between them. For a set of input locations $\mathbf{x}=\{x_1,\ldots,x_n\}$, the covariance elements can then be described by a covariance matrix
\begin{equation}
\mathbf{K}_{\theta}(\mathbf{x}, \mathbf{x})=\left(\begin{array}{cccc}{k_{\theta}\left(x_{1}, x_{1}\right)} & {k_{\theta}\left(x_{1}, x_{2}\right)} & {\dots} & {k_{\theta}\left(x_{1}, x_{n}\right)} \\ {k_{\theta}\left(x_{2}, x_{1}\right)} & {k_{\theta}\left(x_{2}, x_{2}\right)} & {\dots} & {k_{\theta}\left(x_{2}, x_{n}\right)} \\ {\vdots} & {\vdots} & {\vdots} & {\vdots} \\ {k_{\theta}\left(x_{n}, x_{1}\right)} & {k_{\theta}\left(x_{n}, x_{2}\right)} & {\dots} & {k_{\theta}\left(x_{n}, x_{n}\right)}\end{array}\right).
\end{equation}
One can see that the function evaluations at the input points in $\mathbf{x}$ are a sample from a multivariate Gaussian distribution,
\begin{equation}
p(\mathbf{y}(\mathbf{x}))=\mathcal{N}(\boldsymbol{\mu}(\mathbf{x}), \mathbf{K}_{\theta}(\mathbf{x}, \mathbf{x})).
\end{equation}
Here $\mathbf{y}=\left\{y_{1}, y_{2}, \ldots, y_{n}\right\}$ are dependent function values evaluated at input locations $x_{1}, x_{2}, \ldots, x_{n}$, and $\boldsymbol{\mu}$ is a vector of mean function values evaluated at $x_{1}, x_{2}, \ldots, x_{n}$.

In most situations, especially in the context of time-series analysis, our observations are data corrupted by a noise process. We can take account of this by defining
\begin{equation}
y=f(x)+\eta,
\end{equation}
in which $\eta$ denotes the noise item. In common practice, $\eta$ is assumed to be Gaussian distributed $\eta\sim \mathcal{N}\left(0, \sigma_n^{2}\right)$, where $\sigma_n^{2}$ denotes the variance of the noise. For noisy observations, the form of the covariance matrix becomes
\begin{equation}\label{eqn:V}
\mathbf{V}_{\theta}(\mathbf{x}, \mathbf{x})=\mathbf{K}_{\theta}(\mathbf{x}, \mathbf{x})+\sigma_n^{2} \mathbf{I}
\end{equation}
where $\mathbf{I}$ is the identity matrix.

The form of the kernel function together with its hyper-parameter $\theta$ has a great impact on the efficacy of GP approximation. The most commonly used kernel function is arguably the squared exponential (SE) function, given by
\begin{equation}\label{eq:se}
k_{\theta}\left(x_{i}, x_{j}\right)=h^{2} \exp \left[-\left(\frac{x_{i}-x_{j}}{\lambda}\right)^{2}\right],
\end{equation}
in which $\theta\triangleq[h,\lambda]$ is the hyper-parameter. In this paper, we adopt the Matern $5/2$ kernel function, defined as
\begin{equation}\label{eqn:matern}
k_{\theta}\left(x_{i}, x_{j}\right)=\sigma_{f}^{2}\left(1+\frac{\sqrt{5} r}{\sigma_{l}}+\frac{5 r^{2}}{3 \sigma_{l}^{2}}\right) \exp \left(-\frac{\sqrt{5} r}{\sigma_{l}}\right),
\end{equation}
where $r=\sqrt{\left(x_{i}-x_{j}\right)^{T}\left(x_{i}-x_{j}\right)}$ is the Euclidean distance between $x_{i}$ and $x_{j}$.

The SE kernel function is infinitely differentiable and thus may yield unrealistic results for physical processes \cite{stein2012interpolation}. In contrast with the SE kernel function, the Matern class kernel function is better for capturing temporal structures in physical processes due to its finite differentiability \cite{stein2012interpolation}. Specifically, the Matern $5/2$ kernel function is twice differentiable and has been widely used in GP \cite{williams2006gaussian}.
Now we have $\theta\triangleq[\sigma_{f},\sigma_{l}]$. The hyper-parameter of the kernel function describes the general properties of our function \cite{williams2006gaussian}. As shown in Eqn. (\ref{eqn:matern}), $\sigma_{f}$ governs the output scale of our function, $\sigma_{l}$ determines the input scale, and thus the smoothness of our function. For other types of kernel functions used for GP regression, see \cite{williams2006gaussian}.

Denote $\epsilon\triangleq\{\theta,\sigma_n\}$ as the hyper-parameter of the GP model. Then, given an observed dataset $\{\mathbf{x},\mathbf{y}\}$, the value of $\epsilon$ can be determined by maximizing the log marginal likelihood \cite{williams2006gaussian}:
\begin{eqnarray}\label{eqn:mlik}
\log p(\mathbf{y}|\mathbf{x})&=&-\frac{1}{2} \mathbf{y}^{\top}\left(\mathbf{K}_{\theta}(\mathbf{x}, \mathbf{x})+\sigma_n^{2} \mathbf{I}\right)^{-1} \mathbf{y}\\\nonumber
& &-\frac{1}{2} \log \left|\mathbf{K}_{\theta}(\mathbf{x}, \mathbf{x})+\sigma_n^{2} \mathbf{I}\right|-\frac{n}{2} \log 2 \pi.
\end{eqnarray}
A conjugate gradient descent optimization algorithm included in the GPML toolbox \cite{rasmussen2010gaussian} is often used to address the above maximization problem.

Now let consider how to predict the observation $y_{*}$ at a test input location $x_{*}$ based on an observed dataset $\{\mathbf{x},\mathbf{y}\}$ (i.e., the training dataset in the ML jargon). According to the definition of GP, $\mathbf{y}$ and $y_{\star}$ are jointly distributed as follows
\begin{equation}\label{eqn:joint}
p\left(\left[\begin{array}{l}{\mathbf{y}} \\ {y_{*}}\end{array}\right]\right)=\mathcal{N}\left(\left[\begin{array}{c}{\boldsymbol{\mu}(\mathbf{x})} \\ {\mu\left(x_{*}\right)}\end{array}\right],\left[\begin{array}{cc}{\mathbf{K}_{\theta}(\mathbf{x}, \mathbf{x})} & {\mathbf{K}_{\theta}\left(\mathbf{x}, x_{*}\right)} \\ {\mathbf{K}_{\theta}\left(x_{*}, \mathbf{x}\right)} & {k_{\theta}\left(x_{*}, x_{*}\right)}\end{array}\right]\right)
\end{equation}
where $\mathbf{K}_{\theta}\left(\mathbf{x}, x_{*}\right)=[k_{\theta}(x_1, x_{*}) \ldots k_{\theta}(x_n, x_{*})]^{\top}$ and $\mathbf{K}_{\theta}\left(x_{*},\mathbf{x}\right)$ is the transpose of $\mathbf{K}_{\theta}\left(\mathbf{x}, x_{*}\right)$. Using some linear algebra operations, one can derive the posterior distribution of $y_{*}$, which is Gaussian with mean
\begin{equation}\label{eqn:mean}
m_{*}=\mu\left(x_{*}\right)+\mathbf{K}_{\theta}\left(x_{*}, \mathbf{x}\right) \mathbf{K}_{\theta}(\mathbf{x}, \mathbf{x})^{-1}(\mathbf{y}-\boldsymbol{\mu}(\mathbf{x}))
\end{equation}
and variance
\begin{equation}\label{eqn:variance}
\sigma_{*}^{2}=k_{\theta}\left(x_{*}, x_{*}\right)-\mathbf{K}_{\theta}\left(x_{*}, \mathbf{x}\right) \mathbf{K}_{\theta}(\mathbf{x}, \mathbf{x})^{-1} \mathbf{K}_{\theta}\left(\mathbf{x}, x_{*}\right).
\end{equation}
To take account of the observation noise, we can simply substitute the $\mathbf{K}_{\theta}(\mathbf{x}, \mathbf{x})$ term from Eqns. (\ref{eqn:joint})-(\ref{eqn:variance}) with the $\mathbf{V}_{\theta}(\mathbf{x}, \mathbf{x})$ in Eqn. (\ref{eqn:V}).
\subsection{The GPTS model}\label{sec:gpts}
Let consider a time-series $\{t, y_t\}, t=1,2,\ldots$, in which $t\in\mathbb{N}$ denotes the time index, $y_t$ the data observed at $t$. At each time step $t$, we are interested in calculating the predictive distribution of $y_{t+1}$ given all the observations up to time $t$, namely $p(y_{t+1}|y_{1:t})$. Here, $y_{1:t}\triangleq\{y_1,\ldots,y_t\}$. In real practice, a time window can be used to save computation cost or account for non-stationarity of the system. Then the target predictive distribution becomes $p(y_{t+1}|y_{t-\tau+1:t})$, where $\tau\in[1,\ldots,t]$ is the length of the time window.

Now we adapt the GP model into the context of time-series. We describe time-series observations using the model of the form
\begin{equation}
y_{t}=f(t)+\eta_{t}, f \sim \mathcal{GP}\left(\mu, k_{\theta}\right), \eta_{t} \sim \mathcal{N}\left(0, \sigma_n^{2}\right),
\end{equation}
where the time index $t$ is taken as the input (namely the $x$ term in Eqns.(\ref{eqn:gp_def})-(\ref{eqn:variance})), while the observation $y_t$ is the output.
Then, given an observed dataset $\{\mathbf{t},\mathbf{y}\}$, in which $\mathbf{t}=\{t-\tau+1,\ldots,t\}$, $\mathbf{y}=\{y_{t-\tau+1},\ldots,y_t\}$, the predictive distribution of $y_{t+1}$ is given by the mean
\begin{equation}\label{eqn:mean_gpts}
m_{t+1}=\mu\left(t+1\right)+\mathbf{K}_{\theta}\left(t+1, \mathbf{t}\right) \mathbf{V}_{\theta}(\mathbf{t}, \mathbf{t})^{-1}(\mathbf{y}-\boldsymbol{\mu}(\mathbf{t}))
\end{equation}
and the variance
\begin{equation}\label{eqn:variance_gpts}
\sigma_{t+1}^{2}=k_{\theta}\left(t+1, t+1\right)-\mathbf{K}_{\theta}\left(t+1, \mathbf{t}\right) \mathbf{V}_{\theta}(\mathbf{t}, \mathbf{t})^{-1} \mathbf{K}_{\theta}\left(\mathbf{t}, t+1\right).
\end{equation}

This GPTS model generalizes classical time-series models, e.g., autoregressive (AR), autoregressive moving average (ARMA), and Kalman filter \cite{murray2001gaussian,saatcci2010gaussian,turner2012gaussian}.
\section{The proposed INTEL algorithm}\label{sec:INTEL}
In this section, we present details about our INTEL algorithm, which is developed for SOP in the presence of outliers and change points.
\subsection{GPTS Mixture for capturing complex temporal structure}\label{sec:mix_gpts}
\begin{figure}[!htb]
\centering
\includegraphics[width=1.8in,height=1.2in]{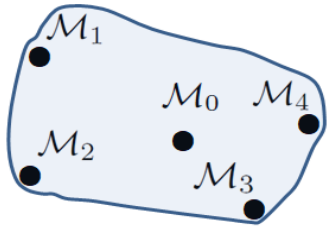}
\caption{A schematic diagram on the modeling capacity of a GP mixture compared with that of a single GP. A black dot denotes the modeling capacity of a single GP. The modeling capacity of the GP mixture is correspondingly a plane spanned by the black dots.}\label{fig:capacity}
\end{figure}

The main idea underlying our algorithm is as follows. We treat a data stream in which outliers and change points are present as a function $f$ with time-varying temporal structures. We use GPTS models to capture temporal structures of $f$. In a GPTS model, each hyper-parameter describes one aspect of the temporal structure underlying the data. For example, for a model with a Matern 5/2 kernel function, the hyper-parameter $\sigma_f$ describes the amplitude of the function, $\sigma_l$ determines its smoothness and $\sigma_n^2$ represents the variance of the observation noise. Given specific values of its hyper-parameters, a GPTS model can capture a specific temporal structure of the data. Suppose that a historical dataset $\{\mathbf{t}_0, \mathbf{y}_0\}$ is pre-available, then we can use a template model $\mathcal{M}_0$ to capture the temporal structure underlying these data.
In most situations, it can be reasonably assumed that a relatively small number of historical data points are pre-available. Since $\mathcal{M}_0$ is obtained based on a very limited number of historical data, its modeling capacity shall be very limited too. That means only using $\mathcal{M}_0$ is impossible to capture temporal structures underlying future data points since non-stationarity is the basic feature of time-series data.
We come up with an idea to enlarge the modeling capacity of $\mathcal{M}_0$. Given $\mathcal{M}_0$ with hyper-parameters $\sigma_{f,0},\sigma_{l,0},\sigma_{n,0}$, we construct a set of candidate models $\mathcal{M}_1,\ldots,\mathcal{M}_M$ based on $\mathcal{M}_0$. That says these candidate models are variants of $\mathcal{M}_0$.
We let all variants share the same mean function with $\mathcal{M}_0$, but take different hyper-parameter values. Note that the term temporal structure used here is defined with hyper-parameters $\sigma_{f},\sigma_{l},\sigma_{n}$, and is not related to the mean function. We use a weighted mixture of these models to cover temporal structures underlying unseen data in future. Although only a limited number of GPTS models can be used, the number of their combinations, defined by their weights, is infinite. It means that the modeling capacity of this mixture model can be much larger than that of $\mathcal{M}_0$, as conceptually illustrated in Figure \ref{fig:capacity}. Hence, we may get much better SOP result based on this mixture model, while, to make the above idea work, we need to answer two questions at first, namely, how to build up the variant models and how to combine all models in an appropriate way to capture the true temporal structure underlying the data. We propose specific techniques to answer the above questions.

Let take an example to show how to construct variants of $\mathcal{M}_0$. Suppose that, given $\mathcal{M}_0$ at the beginning, we believe that the input scale will decrease later, namely, our function will become rougher later. Then we can translate the above belief by introducing a variant model $\mathcal{M}_i, i>0$, for which we assign a smaller input scale value, say $\sigma_{l,i}=0.2\sigma_{l,0}$. The coefficient $0.2$ is related to the limit of the input scale we expect. If we are uncertain whether the input scale will decrease or increase, then except $\mathcal{M}_i$, we can introduce another variant model $\mathcal{M}_j, j>0, j\neq i$, for which we use a bigger input scale value, say $\sigma_{l,j}=5\sigma_{l,0}$.

Notice that for a time-series, abrupt changes in the temporal structure only appear at locations of outliers and change points. Therefore the temporal structure information learned from the historical data can provide important clues for guessing temporal structures in the future data. Our method makes use of such clue information by constructing candidate models on the basis of $\mathcal{M}_0$ and thus is much better than brute force methods that construct candidate models arbitrarily from scratch. The proposed mechanism to handle outliers and change points is deferred to subsection \ref{sec:train_set}.

For the issue of model combination, we treat each model as a hypothesis. Each model is associated with a weight, which represents the probability that this hypothesis is true. The weights of the models are tuned over time to let the weighted mixture of these models capture the non-stationary temporal structure of the data. We resort to a dynamic version of the Bayesian model averaging (DMA) technique to tune the model weights. For more details on the DMA method, see \cite{dai2016robust,liu2011instantaneous,liu2017robust}. Suppose that, at time step $t$, the model $\mathcal{M}_i$ has a weight $\omega_{i,t}>0$, $i\in\{0,1,\ldots,M\}$, $\sum_{i=0}^M\omega_{i,t}=1$. Then the predictive weights of the models at time step $t+1$ are calculated as follows
\begin{equation}\label{eqn:pred_weight}
\hat{\omega}_{i,t+1}=\frac{\omega_{i,t}^{\alpha}}{\sum_{j=0}^{M}\omega_{j,t}^{\alpha}}, i=0,\ldots,M,
\end{equation}
where $0<\alpha<1$ is termed the forgetting parameter. Upon the arrival of the observation $y_{t+1}$, the model weights are updated according to Bayesian formalism as follows
\begin{equation}\label{eqn:weight}
\omega_{i,t+1}=\frac{\hat{\omega}_{i,t+1} p\left(y_{t+1} |\mathcal{M}_i \right)}{\sum_{j=0}^{M}\hat{\omega}_{j,t+1} p\left(y_{t+1} |\mathcal{M}_j\right)}, i=0,\ldots,M,
\end{equation}
where $p\left(y_{t+1} |\mathcal{M}_i \right)$ denotes the likelihood of $y_{t+1}$ under the hypothesis $\mathcal{M}_i, i=0, \ldots, M$.
\subsection{Fusion of GPTS predictions}\label{sec:fusion}
Now we consider how to combine predictions provided by $\mathcal{M}_0, \mathcal{M}_1, \ldots, \mathcal{M}_M$, to yield a fused prediction. Recall that prediction with a single GPTS model is presented in Section \ref{sec:gpts}. Following the setting in Section \ref{sec:gpts}, we focus on the calculation of the predictive distribution of $y_{t+1}$, namely $p(y_{t+1}|y_{t-\tau+1:t})$ (or $p(y_{t+1})$ for short). Denote the predictive distribution of $y_{t+1}$ corresponding to $\mathcal{M}_i$ as $p_i(y_{t+1}|y_{t-\tau+1:t})$ (or $p_i(y_{t+1})$ for short). The mean and the variance of $p_i(y_{t+1})$, denoted as $m_{i,t+1}$ and $\sigma_{i,t+1}^{2}$, are calculated using Eqns.(\ref{eqn:mean_gpts})-(\ref{eqn:variance_gpts}). To calculate $p(y_{t+1})$ based on $p_i(y_{t+1})$, $i=0,\ldots,M$, the POE model of \cite{hinton2002training} can be used. Given multiple probability densities, $p_i(y_{t+1})$, $i=0,\ldots,M$, the POE model describes the target distribution $p(y_{t+1})$ as follows,
\begin{equation}\label{eqn:poe}
p(y_{t+1})=\frac{1}{Z} \Pi_{i=0}^M p_i(y_{t+1}),
\end{equation}
in which $Z$ is a normalizing constant that makes $p(y_{t+1})$ a probability distribution that integrates to 1. Since $p_i(y_{t+1})$, $i=0, \ldots, M$, are all Gaussian, $p(y_{t+1})$ calculated with Eqn.(\ref{eqn:poe}) is still Gaussian, with its mean and variance given by \cite{hinton2002training}
\begin{eqnarray}
m_{t+1}&=&\left(\sum_{i=0}^M(m_{i,t+1}P_i)\right)\left(\sum_{i=0}^MP_i\right)^{-1},\\
\sigma_{t+1}^{2} &=&\left(\sum_{i=0}^MP_i\right)^{-1},
\end{eqnarray}
where $P_i=\left(\sigma_{i,t+1}^{2}\right)^{-1}$.
One can see that the information of the model weights is not involved in the above calculation. In fact, in the original POE model, all models are treated to be equally weighted. Here we generalize the POE model to incorporate the information of the model weights by letting
\begin{equation}\label{eqn:generalized_poe}
p(y_{t+1})\propto \Pi_{i=0}^M \left(p_i(y_{t+1})\right)^{\hat{\omega}_{i,t+1}}.
\end{equation}
Note that, here we use $\hat{\omega}_{i,t+1}$ other than $\omega_{i,t+1}$. This is because the calculation of $\omega_{i,t+1}$ requires access to $y_{t+1}$, see Eqn. (\ref{eqn:weight}). However, the calculation of the predictive density of $y_{t+1}$ is performed at time step $t$. At that time, the real observation $y_{t+1}$ is not accessible. So $\hat{\omega}_{i,t+1}$ is used instead of $\omega_{i,t+1}$ in Eqn.(\ref{eqn:generalized_poe}).
Since $p_i(y_{t+1})$, $i=0, \ldots, M$, are Gaussian, $p(y_{t+1})$ calculated with Eqn. (\ref{eqn:generalized_poe}) is still Gaussian, with its mean and variance given by \cite{cao2014generalized}
\begin{eqnarray}\label{eqn:mean_var}
m_{t+1} &=& \frac{\sum_{i=0}^M(m_{i,t+1}\hat{\omega}_{i,t+1}P_{i})}{\sum_{i=0}^M(\hat{\omega}_{i,t+1}P_{i})}, \\
\sigma_{t+1}^{2}&=&\left(\sum_{i=0}^M(\hat{\omega}_{i,t+1}P_{i})\right)^{-1}.
\end{eqnarray}
The mean $m_{t+1}$ is taken as the prediction of $y_{t+1}$ made at time step $t$. A confidence interval associated with this prediction is also available. For example, a 99.75\% confidence interval is shown to be $[m_{t+1}-3\sigma_{t+1}, m_{t+1}+3\sigma_{t+1}]$.
\subsection{Online outlier detection}
Considering that SOP and online outlier detection are a pair of dual problems, we do outlier detection based on the prediction given by the GPTS mixture mentioned above. Assume that a GPTS mixture model with a rich enough modeling capacity is built up. That says it can produce a reliable prediction for an observation, say $y_{t+1}$, based on the observed data $\{\mathbf{t},\mathbf{y}\}$, provided that $y_{t+1}$ and the elements included in $\mathbf{y}$ are within the same regime. Then outlier detection is performed simply as follows. If the real observation $y_{t+1}$ departs from the confidence interval $[m_{t+1}-3\sigma_{t+1}, m_{t+1}+3\sigma_{t+1}]$, then declare it to be an outlier.
\subsection{Change point detection and instant temporal structure capturing}\label{sec:train_set}
Recall that, in the GPTS model, the predictive distribution of $y_{t+1}$ is calculated based on the training dataset $\{\mathbf{t},\mathbf{y}\}$, in which $\mathbf{t}=\{t-\tau+1,\ldots,t\}$ and $\mathbf{y}=\{y_{t-\tau+1},\ldots,y_t\}$, see Eqns.(\ref{eqn:mean_gpts})-(\ref{eqn:variance_gpts}). This calculation does not take into account the possible presence of outliers or change points in the training dataset. The inclusive of an outlier or a change point will bring detrimental effects to the prediction performance \cite{escalante2005comparison}. To this end, we develop a technique, called adaptive training set formation, to eliminate the negative effects of outliers and change points. We use a potential change point bucket (PCB), denoted as $\{\mathbf{t}', \mathbf{y}'\}$ to save outliers that have been declared consecutively till now. Specifically, if an outlier is declared at $t$, then add $t$ and $y_t$ into $\mathbf{t}'$ and $\mathbf{y}'$, respectively. Otherwise, we empty $\mathbf{t}'$ and $\mathbf{y}'$ and add $t$ and $y_t$ into $\mathbf{t}$ and $\mathbf{y}$, respectively. After that, we check if the number of elements in $\mathbf{t}'$ (or $\mathbf{y}'$) achieves a certain number, say $N$. If so, we declare a change point detection.

Upon a change point is declared, we set $\mathbf{t}=\mathbf{t}'$, $\mathbf{y}=\mathbf{y}'$, and then empty $ \mathbf{t}'$ and $\mathbf{y}'$. In this way, a new training dataset $\{\mathbf{t}, \mathbf{y}\}$ is formed, based on which we can do predictions for future observations in the new regime. To achieve a fast detection of the regime shift, $N$ should take a small value, while, to learn a qualified model to capture the temporal structure of the new regime, the bigger is $N$, the better. We break this dilemma by proposing an instant learning technique with the help of the rich modeling capacity of our GPTS mixture. Recall that in the setting of this paper, a GPTS model is defined by its hyper-parameters $\sigma_f, \sigma_l, \sigma_n$ which describe the temporal structure and the mean function $\mu(\cdot)=C$ that describes the major trend. So constructing a qualified GPTS model is equivalent to finding appropriate values for such hyper-parameters and $C$. If we only have a relatively small number $N$ of labeled data points, it is impossible to find appropriate values for all these parameters. Our idea is to borrow the power of the adaptive weighted mixture model mentioned above to automatically capture the temporal structure of the new regime, while only update the value of $C$ based on these data points, namely
\begin{equation}\label{eqn:mu}
\mu(\cdot)=C=\frac{1}{N}\sum_{i\in\mathbf{t}}y_i.
\end{equation}
See Figure \ref{fig:ec2_cpu_data_pred_init1} for an example performance show of the above mechanism. We can see from the middle panel of Figure \ref{fig:ec2_cpu_data_pred_init1}, when a regime shift appears at $t=2971$, the weight of the previously dominated model $\mathcal{M}_0$ falls rapidly, while, at the same time, the weight of $\mathcal{M}_1$, whose hyper-parameter setting is more fit to the new regime, rises abruptly. The above result clearly shows that the adaptive model weighting mechanism of our method takes effect, rendering our mixture model capture the temporal structure of the new regime instantly.
\subsection{Implementation of the INTEL algorithm}\label{sec:intel}
A pseudo-code to implement the INTEL algorithm is presented in Algorithm \ref{alg:INTEL}. The computational complexity of each major operation is marked. A formal computation complexity analysis is deferred to subsection \ref{sec:compplexity}.
\begin{algorithm}
\caption{The Proposed INTEL Algorithm}
\label{alg:INTEL}
\begin{algorithmic}[1] 
\STATE Input: $N$, $\tau$, $\mu(\cdot)$, $\alpha$, $\omega_{i,0}, \epsilon_i, i=0,\ldots,M$, $L$ (refer to subsection \ref{sec:mix_gpts} for initialization issues about the input).
\STATE $\mathbf{t}\leftarrow\{ \}$, $\mathbf{y}\leftarrow\{ \}$,$\mathbf{t}'\leftarrow\{ \}$, $\mathbf{y}'\leftarrow\{ \}$;
\FOR{$t$=0, 1, \ldots} \FOR{$i=0, 1,\ldots,M$}
\STATE Calculate $m_{i,t+1}, \sigma_{i,t+1}^2$ using Eqns.(\ref{eqn:mean_gpts})-(\ref{eqn:variance_gpts}); ($\mathcal{O}\left(\tau^{3}\right)$)
\ENDFOR
\STATE Calculate $\hat{\omega}_{i,t+1}$ with Eqn.(\ref{eqn:pred_weight}), $i=0,\ldots,M$; ($\mathcal{O}\left(M\right)$)
\STATE Calculate $\omega_{i,t+1}$ with Eqn.(\ref{eqn:weight}), $i=0,\ldots,M$; ($\mathcal{O}\left(M\right)$)
\STATE Calculate $m_{t+1}, \sigma_{t+1}^2$ using Eqns.(\ref{eqn:mean_var})-(22); ($\mathcal{O}\left(M\right)$)
\IF {$y_{t+1}<m_{t+1}+3\sigma_{t+1}$ $\&$ $y_{t+1}>m_{t+1}-3\sigma_{t+1}$}
\STATE Add $t+1$, $y_{t+1}$ into $\mathbf{t}$ and $\mathbf{y}$, respectively;
\STATE $\mathbf{t}'\leftarrow\{ \}$, $\mathbf{y}'\leftarrow\{ \}$;
\IF {the size of $\mathbf{t}$ achieves multiples of $L$}
\STATE Let $\mu(\cdot)=C$, where $C$ equals the average of the last $L$ data items that have been added into $\mathbf{y}$; ($\mathcal{O}\left(L\right)$)
\ENDIF
\ELSE
\STATE Add $t+1$, $y_{t+1}$ into $\mathbf{t}'$ and $\mathbf{y}'$, respectively;
\IF {the size of $\mathbf{t}'$ achieves $N$}
\STATE (Optionally) declare $y_{t+1}$ to be a change point;
\STATE $\mathbf{t}\leftarrow\mathbf{t}'$; $\mathbf{y}\leftarrow\mathbf{y}'$;
\STATE Update $\mu(\cdot)$ with Eqn.(\ref{eqn:mu}); ($\mathcal{O}\left(N\right)$)
\ELSE
\STATE (Optionally) declare $y_{t+1}$ to be an outlier;
\ENDIF
\ENDIF
\IF {$\exists j\in\mathbf{t}, j<t+1-\tau+1$}
\STATE Remove $j$, $y_j$ from $\mathbf{t}$ and $\mathbf{y}$, respectively;
\ENDIF
\STATE Output $m_{t+1}, \sigma_{t+1}^2$.
\ENDFOR
\end{algorithmic}
\end{algorithm}
\subsection{Algorithm initialization}\label{sec:discuss_init}
Given the mean function $\mu(\cdot)$, $\epsilon_0$ is initialized by maximizing the log marginal likelihood based on an observed  dataset $\{\mathbf{t}_0, \mathbf{y}_0\}$. Then $\epsilon_i, i>0$ is specified in a way as presented in the second paragraph of subsection \ref{sec:mix_gpts}.
Note that the efficiency of our algorithm does not depend on a fixed model set, because different model sets can have the same function for one specific dataset. In subsection \ref{sec:test_init}, we present an example case that shows the experimental evidence of the above argument. In that example, the inclusive of low-quality models have little impact on the prediction performance, because the model weighting procedure (see Eqns.(\ref{eqn:pred_weight})-(\ref{eqn:weight})) automatically assigns tiny weights to those low-quality models, and thus eliminates their negative effects.

Now we discuss initialization issues about the other parameters. We specify $\mu(\cdot)$ to be a constant value function, $\mu(\cdot)=C$, where $C$ is initialized to be the average of those data points included in $\mathbf{y}_0$. All model weights are initialized to be $1/(M+1)$. As for $N$, the smaller is its value, the earlier a change point can be detected, while, it can not be arbitrarily small, otherwise, a detected change point may be an outlier. We set its value at 3 to give a balance between the timeliness of change point detection and the discrimination between a change point and outliers. For $\alpha$, we follow our previous work in \cite{liu2017robust}, setting its value at 0.9. The parameter $\tau$ represents the length of the time window, while in our algorithm, it just determines the maximum number of training data points allowed for use in calculating the predictive distribution, see Eqns.(\ref{eqn:mean})-(\ref{eqn:variance}). The actual number of training data points and which data points within the time window will be selected as training data points are both determined by the adaptive training set formation procedure described in subsection \ref{sec:train_set}. That says the value of $\tau$ has much less impact on the prediction performance of our algorithm than for traditional time window based methods. In our experiments, we set $\tau=20$. Lastly, the parameter $L$ controls the period for fine-tuning the mean function. We select to update the mean function periodically, because, even within one regime, the time-series data may still be non-stationary. Fine-tuning the mean function is beneficial for capturing the changes in the trend of our function. In our experiments, we set $L=10$.
\subsection{A formal analysis of the computational complexity of the INTEL algorithm}\label{sec:compplexity}
We mark the computational complexity for each major operation in Algorithm \ref{alg:INTEL}. It shows that the most computationally complex operation is the inversion of a $\tau\times\tau$ matrix in line 5, which scale as $\mathcal{O}\left(\tau^{3}\right)$. The matrix inversion operation is performed $M$ times, leading to a complexity $\mathcal{O}\left(M\tau^{3}\right)$ in total. The calculation in line 7 consists of simple numerical operations that scales as $\mathcal{O}\left(M\right)$. The computation in line 8 involves $M+1$ Gaussian likelihood calculations scaling as $\mathcal{O}\left(M\right)$. The operation in line 9 involves multiplication and addition calculations that scale as $\mathcal{O}\left(M\right)$. The remaining operations include the comparison operation in line 10, the average operations in lines 14 and 21, whose computational complexity is negligible compared with others. To summarize, all operations in the INTEL algorithm are of the linear algebra type and scale as $\mathcal{O}\left(M\tau^{3}\right)$. This algorithm shall be highly computationally efficient if $\tau$ and $M$ take small values.
\section{Connections to relevant works in the literature}\label{sec:connection}
As a GPTS model-based method, our INTEL algorithm is relevant with all existent works that involve the GPTS model and take into account outliers or change points. See e.g., \cite{chandola2011gaussian,saatcci2010gaussian,roberts2013gaussian,osborne2012real,garnett2010sequential,osborne2011machine}, to name just a few. A common feature of these existent works is that they all try to design one accurate complex model to cover all cases that may happen in the future, including the appearance of outliers or change points. For example, the algorithm in \cite{vanhatalo2009gaussian} employs a heavy-tailed student's $t$ observation noise model to take into account the presence of outliers. The fault bucket algorithm in \cite{osborne2011machine} takes account of faulty data with a Gaussian distribution with a very wide variance. The approach in \cite{garnett2010sequential} uses a non-stationary kernel function to take into account the appearance of change points. A price to pay for applying such complex models is a significantly increased complexity in the inference. For example, to use the student's $t$ model of \cite{vanhatalo2009gaussian}, one has to use Gibbs sampling \cite{geweke1993bayesian} or variational methods \cite{tipping2005variational} for inference. To use the model of \cite{garnett2010sequential}, one has to address a complex Bayesian quadrature to calculate the predictive distribution.

Different from the aforementioned methods, neither Monte Carlo sampling nor quadrature operation is required to implement our INTEL algorithm, since all operations are of the linear algebra type, see details in subsection \ref{sec:compplexity}. The INTEL algorithm does not try to design one accurate complex model, but to construct a model set to cover possible temporal structures in future data. Each member in the model set is an inaccurate model, while it captures one type of temporal structure and allows closed-form inference. A dynamic data-driven weighting mechanism is used to combine members in the model set, rendering the resulting mixture of GPTS models owns a rich modeling capacity to cover complex temporal structures that may appear in future data.

Lastly, our method has connections to regime-switching Markov or state-space model based approaches for non-stationary time series \cite{vasas2007two,boys2000detecting,takeuchi2006unifying}, online metric learning methods in e.g., \cite{zhong2017slmoml}, and deep neural networks (DNN) based sequence data representation learning methods \cite{sak2015learning,jurtz2017introduction,jia2019towards,ma2016end,gao2017video}. The link between state-space models and GP is can be found in \cite{solin2014explicit,hartikainen2010kalman,sarkka2013spatiotemporal,reece2010introduction}. The connection between GP and neural networks traces back to Neal's work in \cite{neal1996priors}, which shows that certain types of neural networks with one hidden layer of infinite size are identical to a GP model with a specific type of covariance function. A recent study on relationships between DNN and GP can be found in \cite{lee2017deep}. Despite of the intrinsic theoretical link between GP and DNN, from the application point of view, there are several basic differences between GP and DNN. Specifically, the former belongs to the class of nonparametric modeling approaches, while the latter is usually parametric. The former can produce a point estimate as well as its uncertainty measure, while the latter usually can only yield a point estimate. The former is more attractive for adding side information, a property we adopt here to develop the INTEL algorithm, due to its Bayesian nature; while the latter is more attractive for dealing with larger datasets that own non-local smoothing features in e.g., natural languages and speeches \cite{jia2019towards,sak2015learning}.
\section{Experiments}\label{sec:experiment}
We conducted extensive experiments to evaluate our INTEL algorithm. In subsection \ref{sec:test_init}, we present an experiment conducted to validate the efficacy of the model initialization procedure presented in subsection \ref{sec:mix_gpts}. In subsection \ref{sec:cpd_detect}, we show results about its performance for online outlier and change point detection. A quantitative evaluation of its prediction performance is presented in subsection \ref{sec:pred_comp}. Finally, in subsection \ref{sec:robust_test}, we tested its robustness when working under undesirable cases.
\subsection{An experiment for testing the model initialization procedure}\label{sec:test_init}
\begin{figure}[!htb]
\centering
\includegraphics[width=3.5in,height=1.5in]{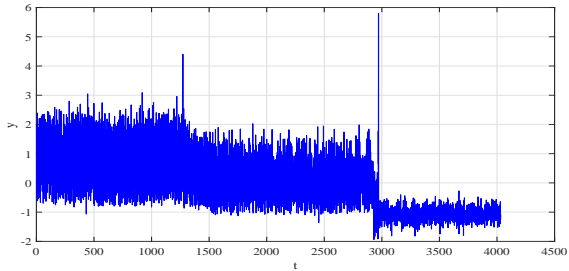}
\caption{A CPU usage dataset collected from a server in Amazon's east coast datacenter \cite{ahmad2017unsupervised}}\label{fig:ec2_cpu_data}
\end{figure}

We check the efficacy of our method for initializing $\mathcal{M}_i$'s, $i>0$, which is presented in subsection \ref{sec:mix_gpts}. We use a CPU usage dataset collected from a server in Amazon's east coast data center \cite{ahmad2017unsupervised}, as shown in Figure \ref{fig:ec2_cpu_data}. In this dataset, a change point appears at around the 3,000th time step (which is exactly the 2,971st time step). After that, both the mean and the output scale of the dataset change significantly. We take the first 200 data points as the historical dataset used for initializing hyper-parameters of $\mathcal{M}_0$. Now let compare two initialization settings for $\mathcal{M}_i, i>0$.

\begin{figure}[!htb]
\centering
\includegraphics[width=3.5in,height=1.5in]{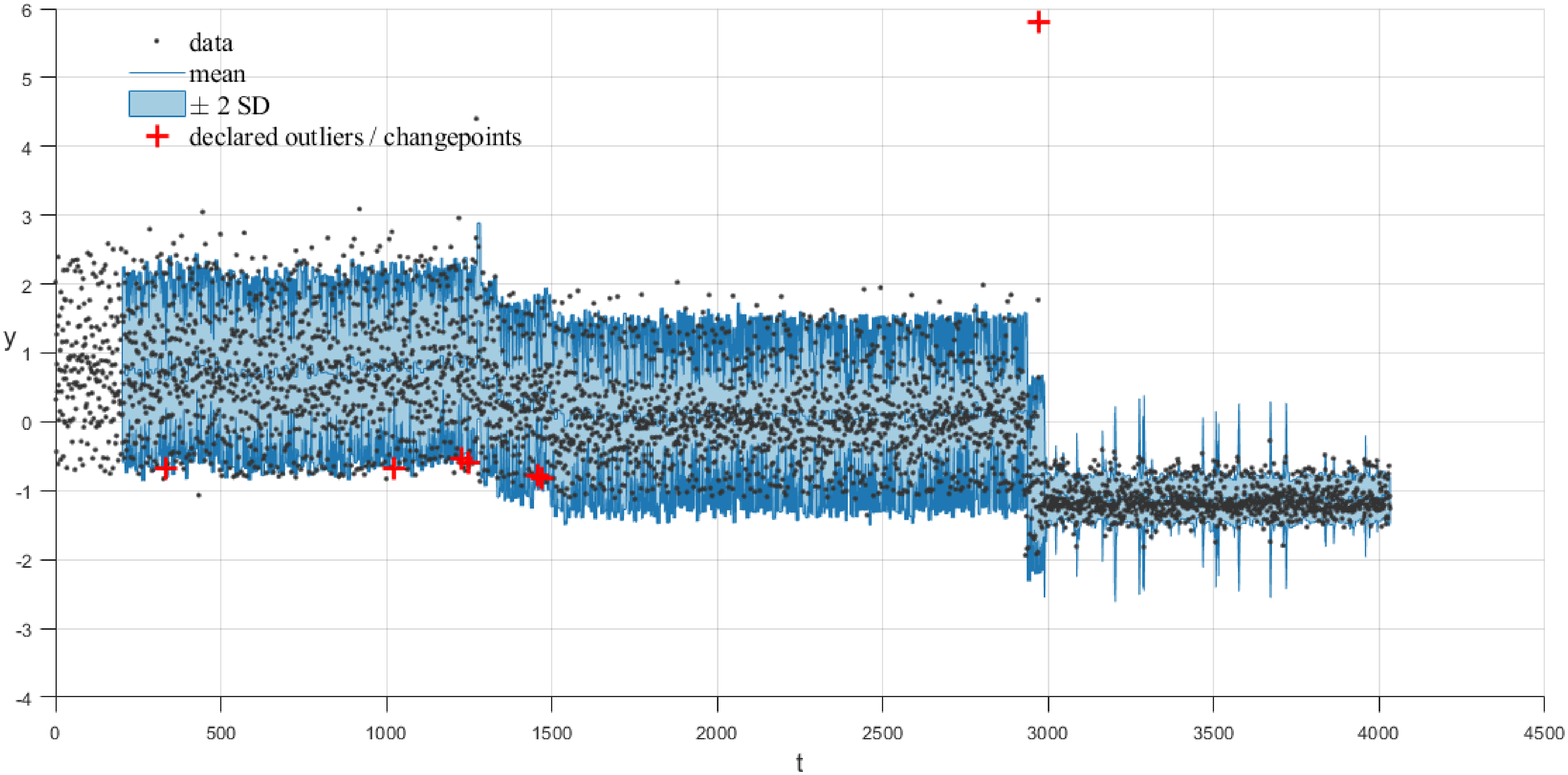}\\
\includegraphics[width=3.5in,height=1.5in]{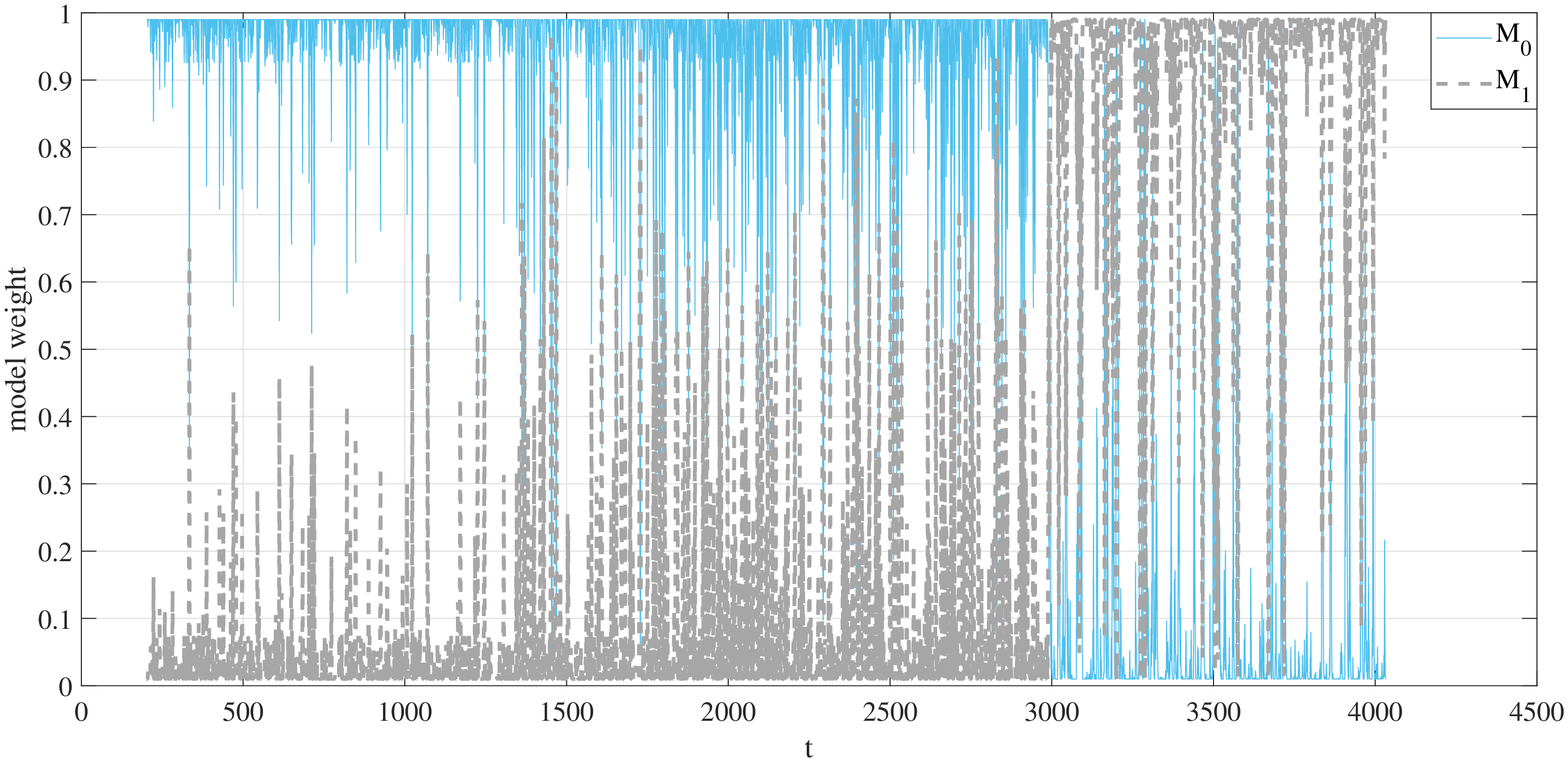}\\
\includegraphics[width=3.5in,height=1.5in]{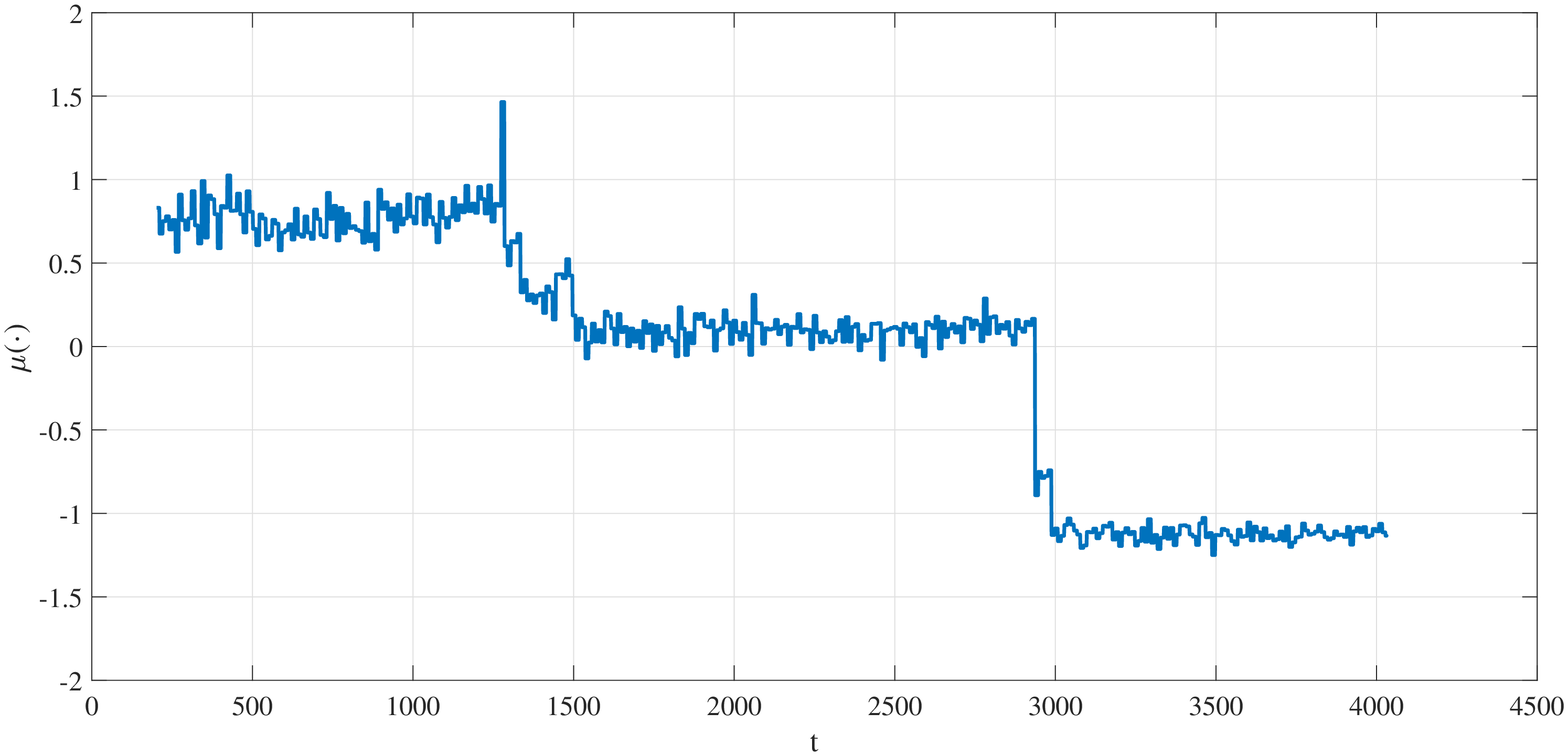}
\caption{The output of the INTEL algorithm for the CPU usage dataset. This result is associated with the first initialization setting for INTEL (see the text in subsection \ref{sec:test_init} for more details). In the upper panel, the symbol SD represents standard derivation, namely the $\sigma$ in Eqn.(22). The middle panel presents the model weights (corresponding to $\omega_{i,t+1}$ in Eqn.(\ref{eqn:weight})). The bottom panel presents the mean function $\mu(\cdot)$, see lines 14 and 21 in Algorithm 1 for its adaptation. The first 200 data points are used for hyper-parameter initialization for $\mathcal{M}_0$.}\label{fig:ec2_cpu_data_pred_init1}
\end{figure}

In the first setting, only one variant of $\mathcal{M}_0$ is used. The hyper-parameter values of $\mathcal{M}_1$ are the same as that of $\mathcal{M}_0$, except that $\sigma_{f,1}=0.2\sigma_{f,0}$. As shown in Figure \ref{fig:ec2_cpu_data}, the real situation is that the output scale of the data decreases markedly after the 2,971st time step. Therefore, the hyper-parameter setting of $\mathcal{M}_1$ is more suitable for the temporal structure of the data after the 2,971th time step. We wonder if our INTEL algorithm can automatically capture this regime shift by increasing the weight of $\mathcal{M}_1$ at that time. The answer is yes, as shown in Figure \ref{fig:ec2_cpu_data_pred_init1}. It is shown that the weight of $\mathcal{M}_1$ rises rapidly, while that of $\mathcal{M}_0$ decreases abruptly, at the 2,971st time step. This result confirms the INTEL algorithm's capability for instant temporal structure learning.

\begin{figure}
\centering
\includegraphics[width=3.5in,height=1.5in]{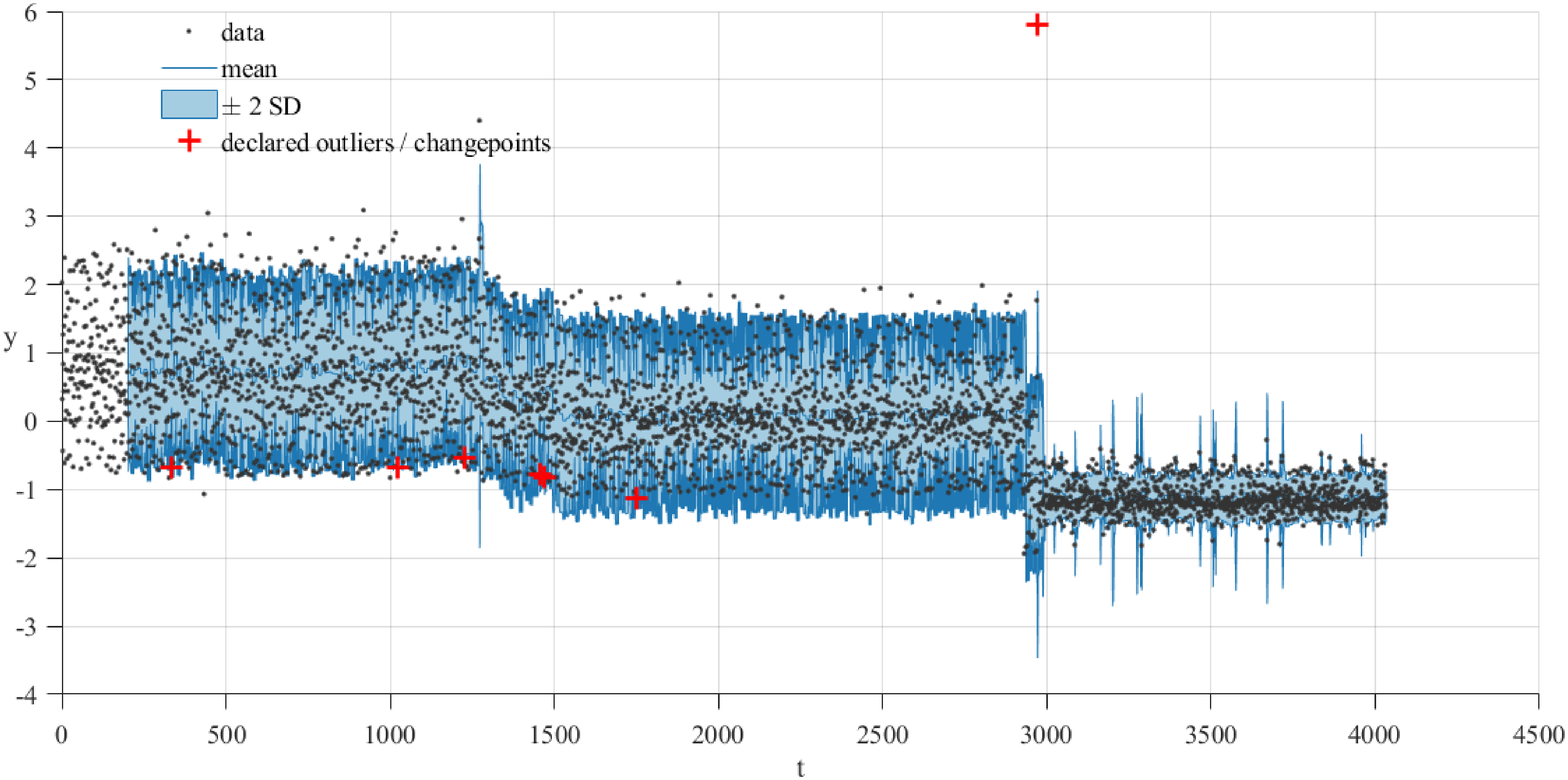}\\
\includegraphics[width=3.5in,height=1.5in]{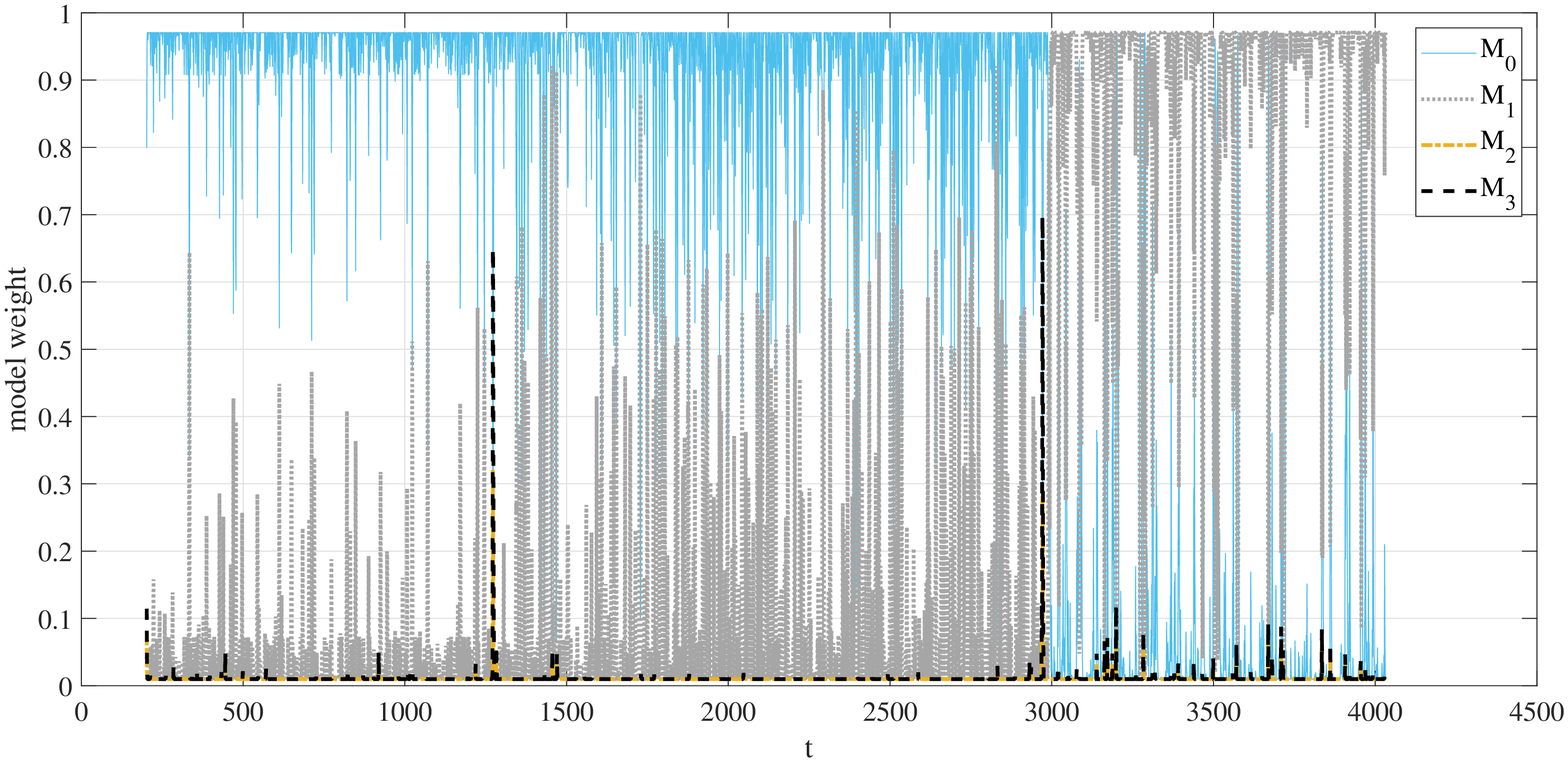}\\
\includegraphics[width=3.5in,height=1.5in]{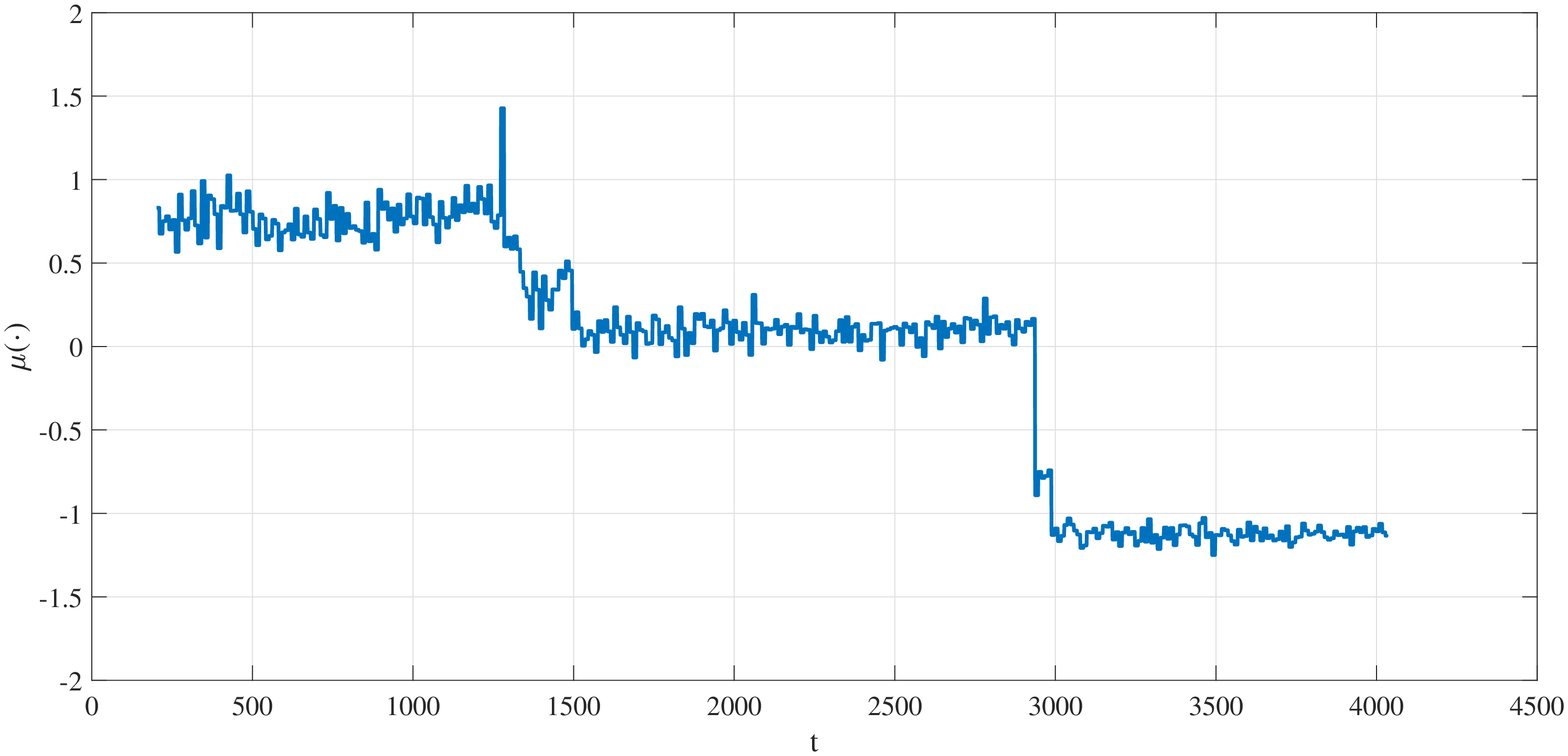}
\caption{The experimental result associated with the second initialization setting (see the text in subsection \ref{sec:test_init} for more details). Compared with Figure \ref{fig:ec2_cpu_data_pred_init1}, two new low-quality models, namely $\mathcal{M}_2$ and $\mathcal{M}_3$, are added into the model set. The middle panel shows that the INTEL algorithm assigns for $\mathcal{M}_2$ and $\mathcal{M}_3$ tiny weight values almost all the time, except at $t=1272, 2971$, where a regime shift happens.}\label{fig:ec2_cpu_data_pred_init2}
\end{figure}

In the second initialization setting, we maintain the same $\mathcal{M}_0$ and $\mathcal{M}_1$ for use as in the first initialization setting, while adding two new low-equality models, $\mathcal{M}_2$ and $\mathcal{M}_3$, for which we set $\sigma_{f,2}=15\sigma_{f,0}$ and $\sigma_{f,3}=10\sigma_{f,0}$, respectively. The values of the other hyper-parameters of $\mathcal{M}_2$ and $\mathcal{M}_3$ are the same as that of $\mathcal{M}_0$ and $\mathcal{M}_1$. Here the goal is to check if the inclusive of such low-quality models can lead to the failure of INTEL. The result is shown in Figure \ref{fig:ec2_cpu_data_pred_init2}. Comparing Figure \ref{fig:ec2_cpu_data_pred_init2} with Figure \ref{fig:ec2_cpu_data_pred_init1}, one can see that after adding the low-quality models, the prediction performance of INTEL is almost unchanged. That says, for this case, the INTEL algorithm is robust to a model set that contains low-quality models. As shown in the middle panel of Figure \ref{fig:ec2_cpu_data_pred_init2}, the reason for this robustness is that the INTEL algorithm only assigns tiny weights to $\mathcal{M}_2$ and $\mathcal{M}_3$ almost all the time, except at $t = 1,272$ and $t=2,971$, where there is an outlier or change point declared.
\subsection{Experiments for online outliers / change points detection}\label{sec:cpd_detect}
Although the central aim of the INTEL algorithm is to do online prediction, we are also interested in whether it can declare outliers and change points in the right way. We used a bunch of real datasets to do the test. We also included the fault bucket algorithm of \cite{osborne2011machine}, which is GPTS model-based, and the BOCPD algorithm of \cite{turner2009adaptive}, which is not GPTS model-based, but Bayesian-based, as benchmark methods for comparison. Except for providing anomaly detections, the fault bucket algorithm and INTEL also do one-step-ahead prediction. Every dataset is pre-processed by a data normalization operation. The normalized dataset has mean zero and standard error 1. For each algorithm, we selected the same portion of the dataset as the historical data used for hyper-parameter initialization. For the INTEL algorithm, we used 8 candidate models, each corresponding to a combination of values for the hyper-parameters $\sigma_f, \sigma_l$ and $\sigma_n$. There are two candidate values for each hyper-parameter, e.g., for $\sigma_f$, one candidate value is $\sigma_{f,0}$, and the other is $r\cdot\sigma_{f,0}$, where the value of $r$ is selected based on prior knowledge. For example, for the CPU usage data case mentioned above in subsection \ref{sec:test_init}, we set $r=0.2$ for $\sigma_f$ to describe our prior knowledge that the observation amplitude will decrease during some period. In subsection \ref{sec:robust_test}, we tested the robustness of INTEL when adopting inaccurate prior knowledge by setting an inappropriate $r$ value.
\subsubsection{Well-log dataset}
\begin{figure}
\centering
\includegraphics[width=3.5in,height=1.5in]{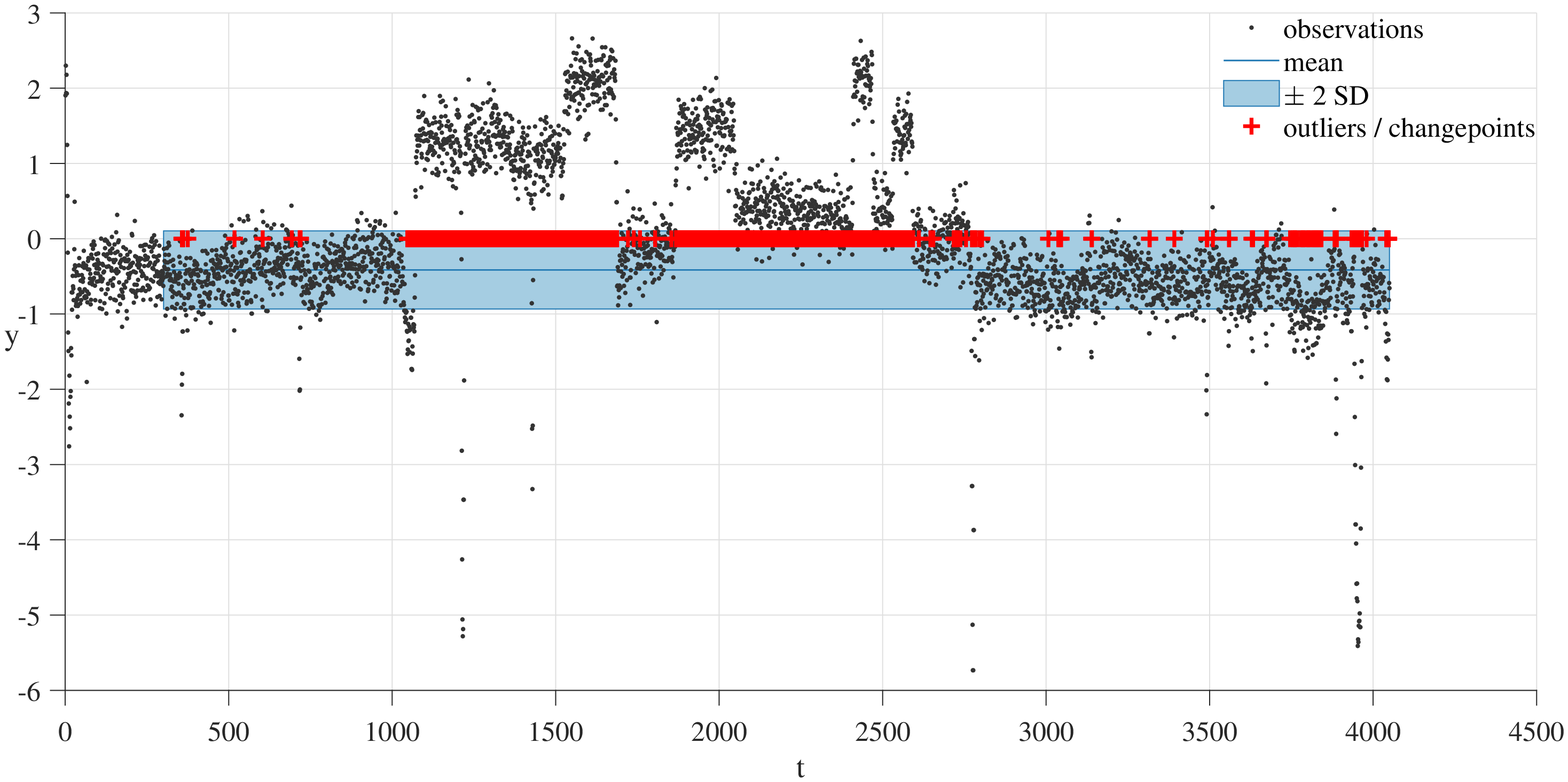}\\
\includegraphics[width=3.5in,height=1.5in]{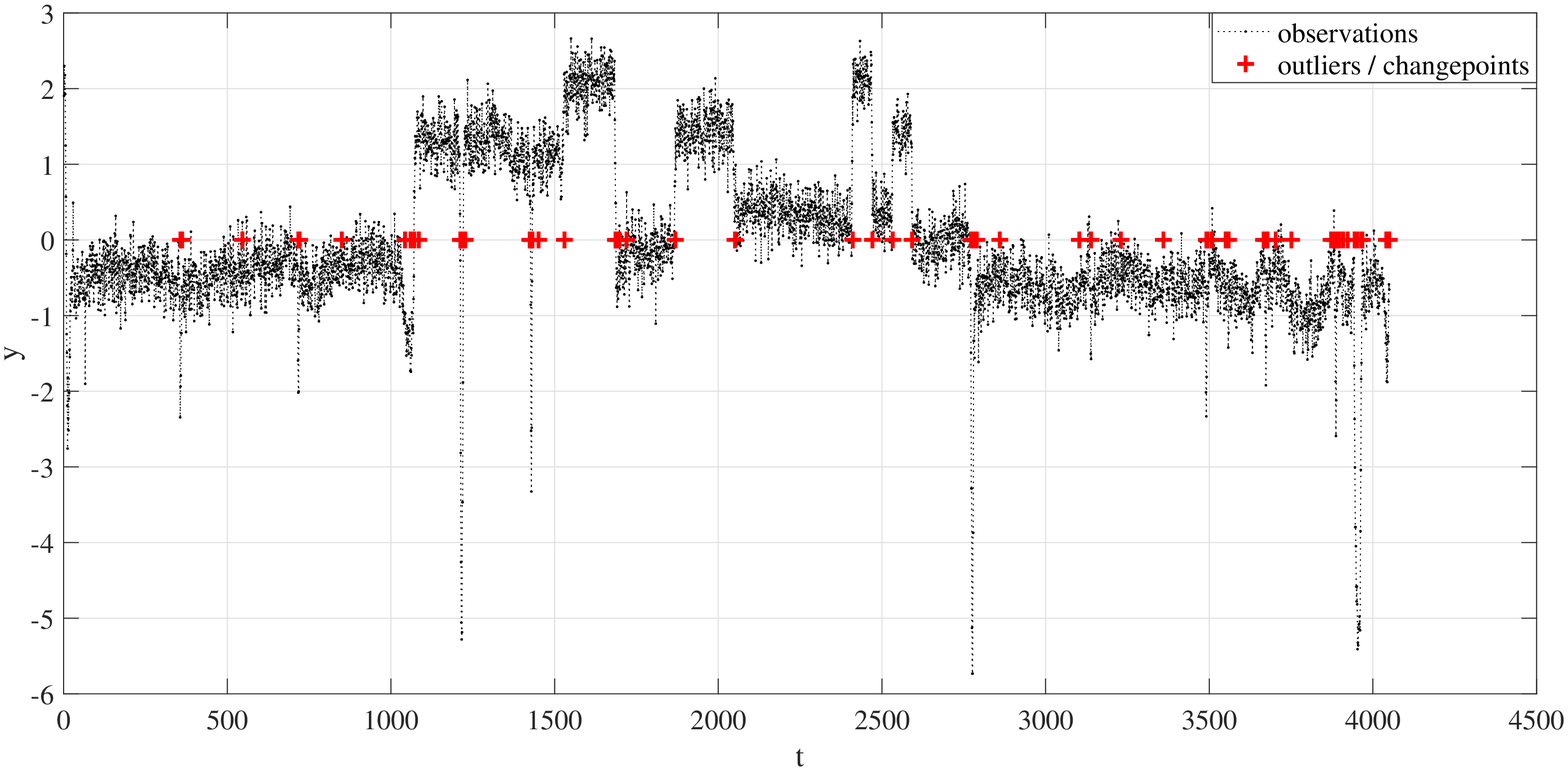}\\
\includegraphics[width=3.5in,height=1.5in]{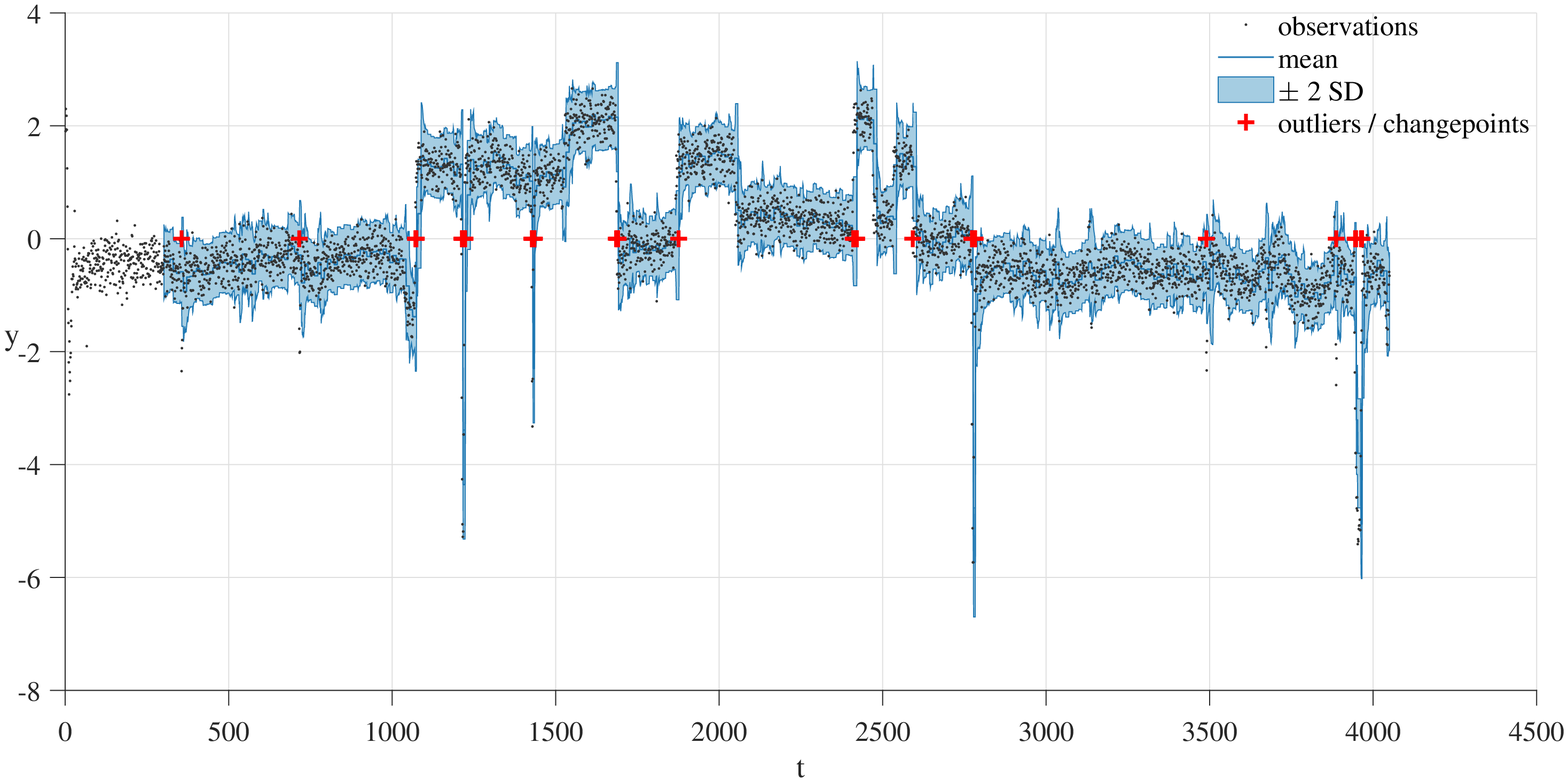}
\caption{Sequential real-time change point detection for the well-log dataset. The upper, the middle and the bottom panels present results corresponding to the fault bucket algorithm \cite{osborne2011machine}, the BOCPD algorithm \cite{turner2009adaptive}, and our proposed INTEL algorithm, respectively. Data points between $t=100$ and $t=300$ are used for hyper-parameter initialization. The algorithm begins to work at $t=301$.}\label{fig:well}
\end{figure}

\begin{figure}
\centering
\includegraphics[width=3.5in,height=1.5in]{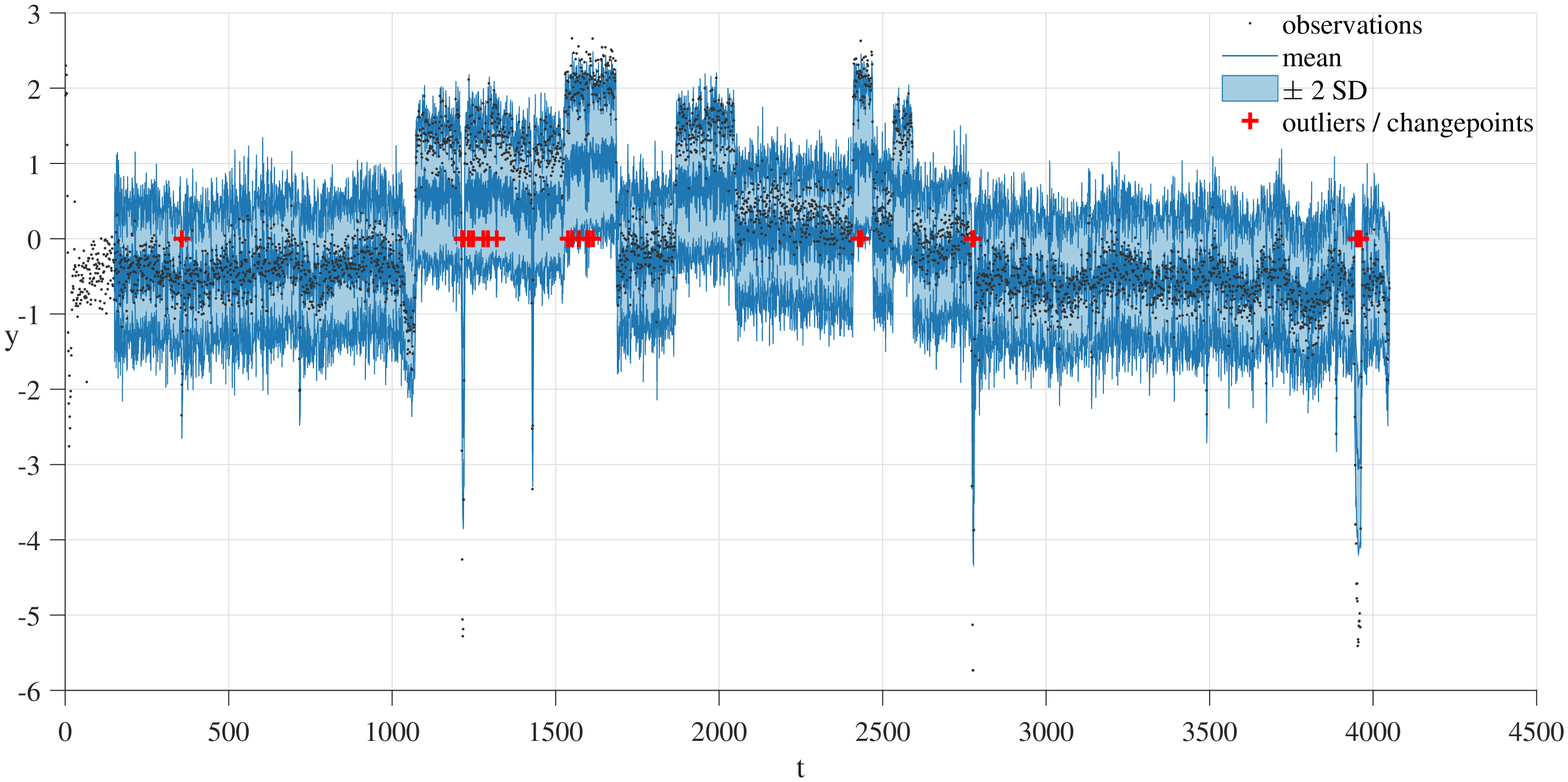}
\caption{Sequential real-time change point detection for the well-log dataset with the fault bucket algorithm \cite{osborne2011machine}. Different from Figure \ref{fig:well}, here the first 150 data points are used for hyper-parameter initialization.}\label{fig:well_fault_bucket2}
\end{figure}

The well-log dataset is widely used in the context of change point detection \cite{turner2009adaptive,turner2012gaussian}. It is a time-series consisting of 4,050 measurements of nuclear magnetic response, which are made during the drilling of a well \cite{garnett2010sequential}. The change here has a clear physical meaning, namely a transition between different strata of rock. The result is plotted in Figure \ref{fig:well}. It is shown that INTEL and BOCPD successfully report all major regime transitions, while the fault bucket algorithm fails to adapt to regime shifts and reports too many false anomaly detections. We checked the reason for the failure of the fault bucket algorithm and found that its performance is highly dependent on the selected data points used for hyper-parameter initialization. If we use the first 150 data points, which contain rougher temporal structures during the first 50 time steps, for hyper-parameter initialization, then the fault bucket algorithm performs much better, as shown in Figure \ref{fig:well_fault_bucket2}. This is due to that the presence of the first 50 data points in the training dataset renders the model capable of capturing rougher temporal structures. Right now the fault bucket algorithm reports much less false anomaly detections, while it has miss-detections. Further, by comparing the bottom panel of Figure \ref{fig:well} with Figure \ref{fig:well_fault_bucket2}, one can see that INTEL is still better than the fault bucket algorithm in terms of prediction performance since the former yields lower-valued variances and thus more certain predictions.
\subsubsection{An ECG dataset}
\begin{figure}
\centering
\includegraphics[width=3.5in,height=1.5in]{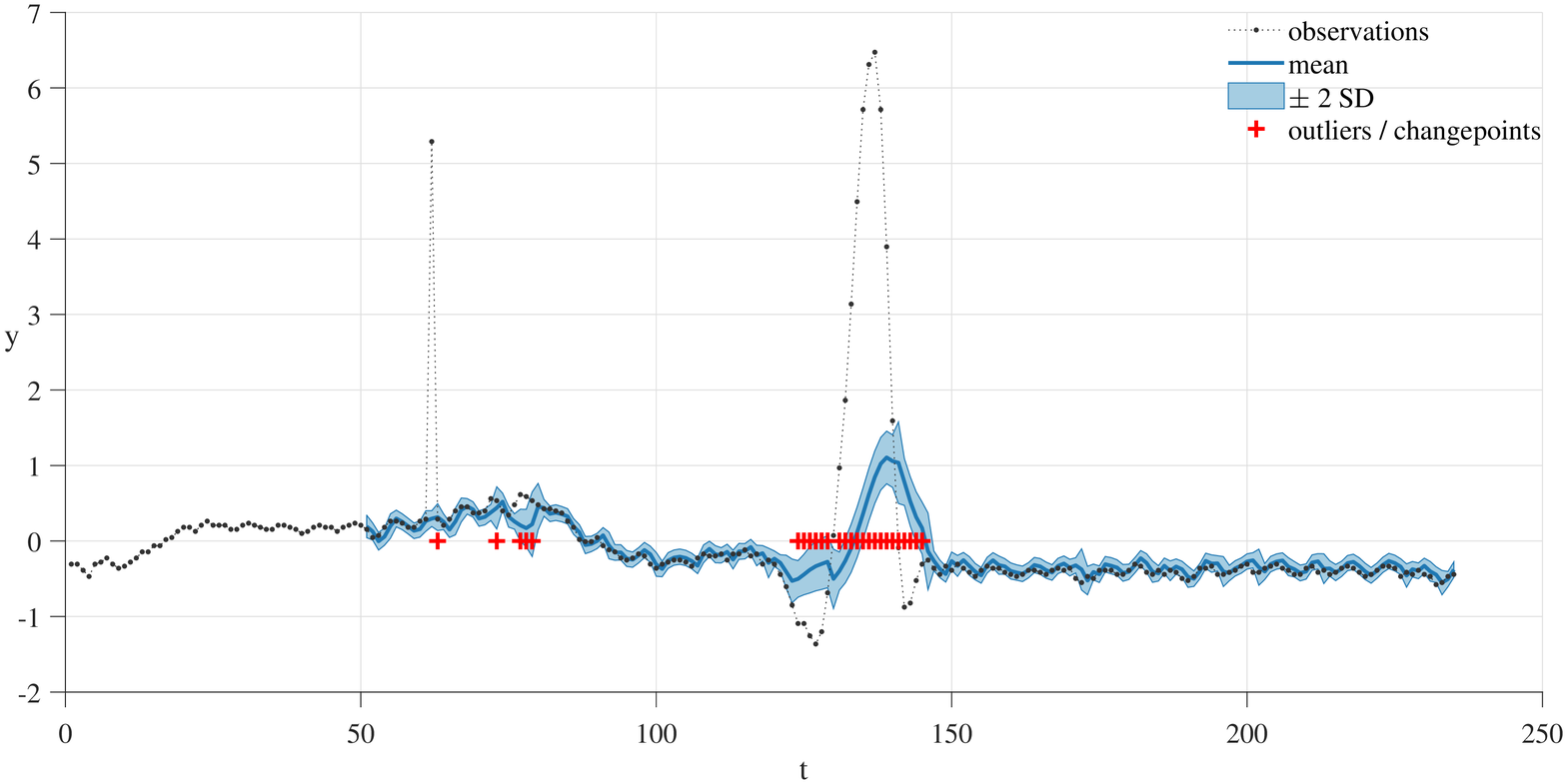}\\
\includegraphics[width=3.5in,height=1.5in]{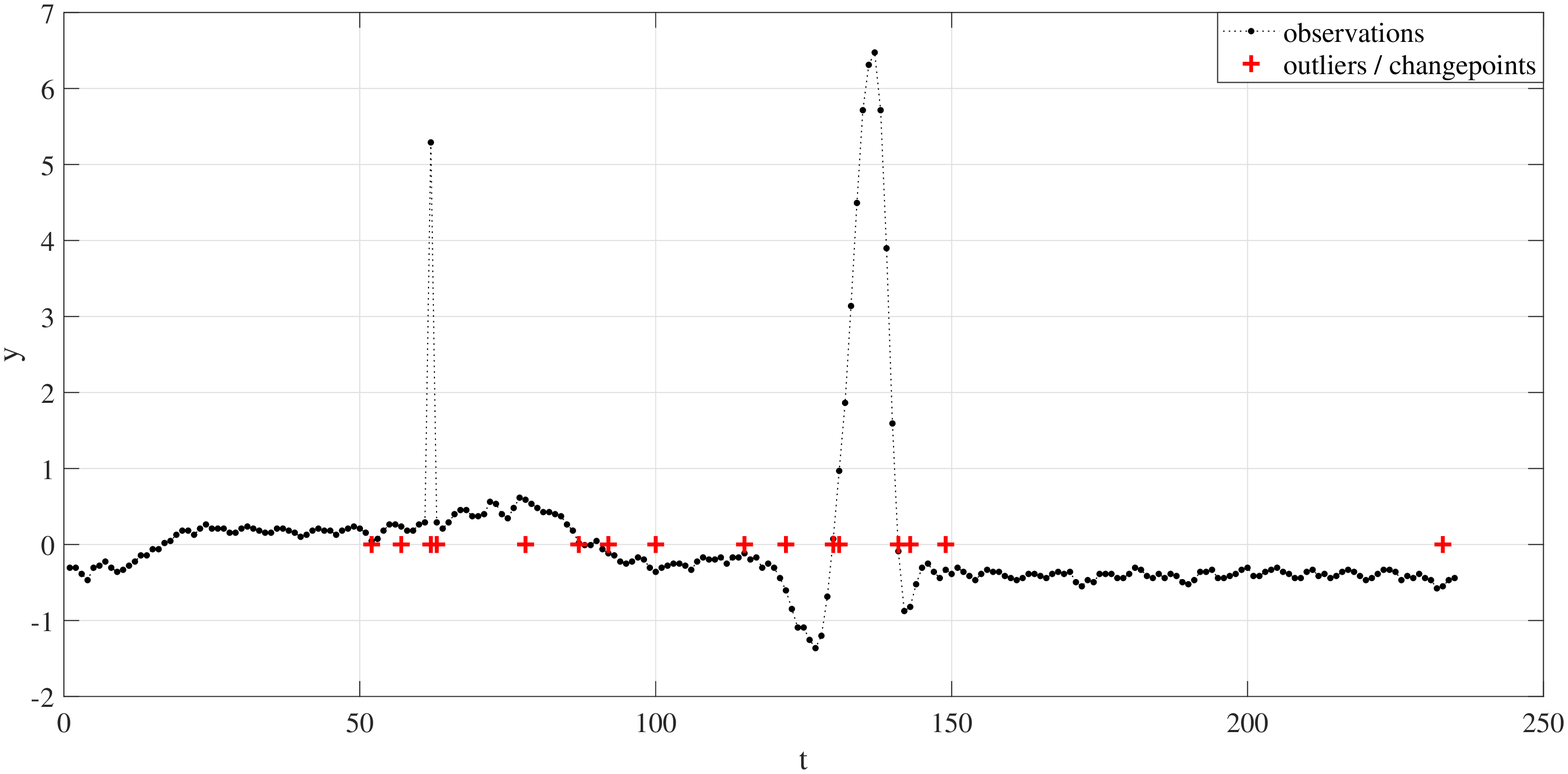}\\
\includegraphics[width=3.5in,height=1.5in]{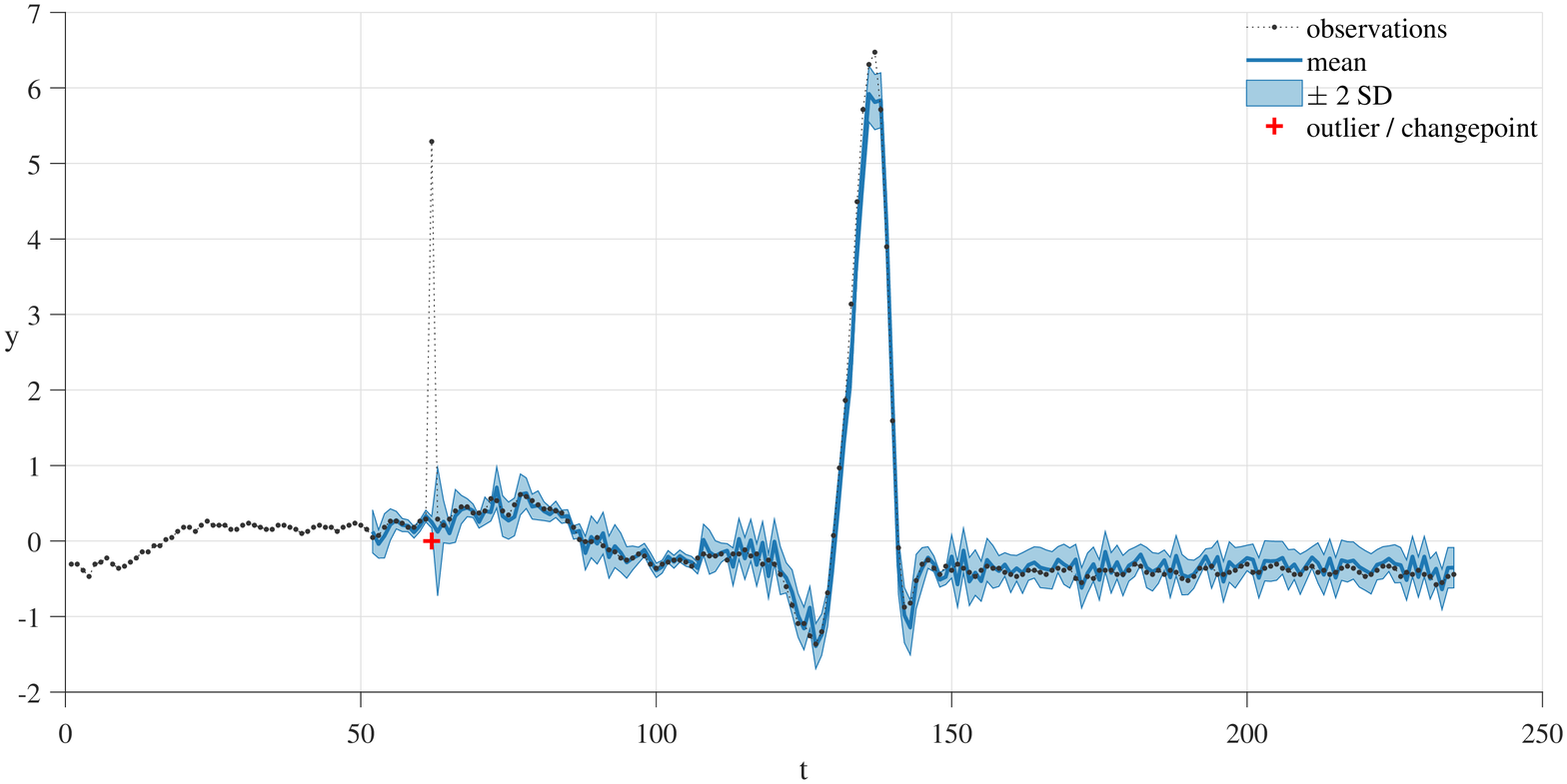}
\caption{Sequential real-time outlier detection for an ECG dataset. The upper, the middle and the bottom panels give results corresponding to the fault bucket algorithm \cite{osborne2011machine}, the BOCPD algorithm \cite{turner2009adaptive}, and our proposed INTEL algorithm, respectively. The first 50 data points are used for hyper-parameter initialization.}\label{fig:ECG}
\end{figure}

This dataset is obtained by injecting an artificial outlier into a piece of real electrocardiogram (ECG) time-series. It comprises 235 observations. The outlier appears at the 62nd time step followed by a saccade that happens from about the 130th time step to about the 145th time step. The one-step-ahead prediction result is shown in Figure \ref{fig:ECG}. One can see that all algorithms considered have successfully detected the true outlier. Our INTEL algorithm is shown to be robust to both the outlier and the saccade, while the fault bucket algorithm fails to yield accurate predictions during the saccade period.
\subsubsection{A Numenta benchmark data}
\begin{figure}
\centering
\includegraphics[width=3.5in,height=1.5in]{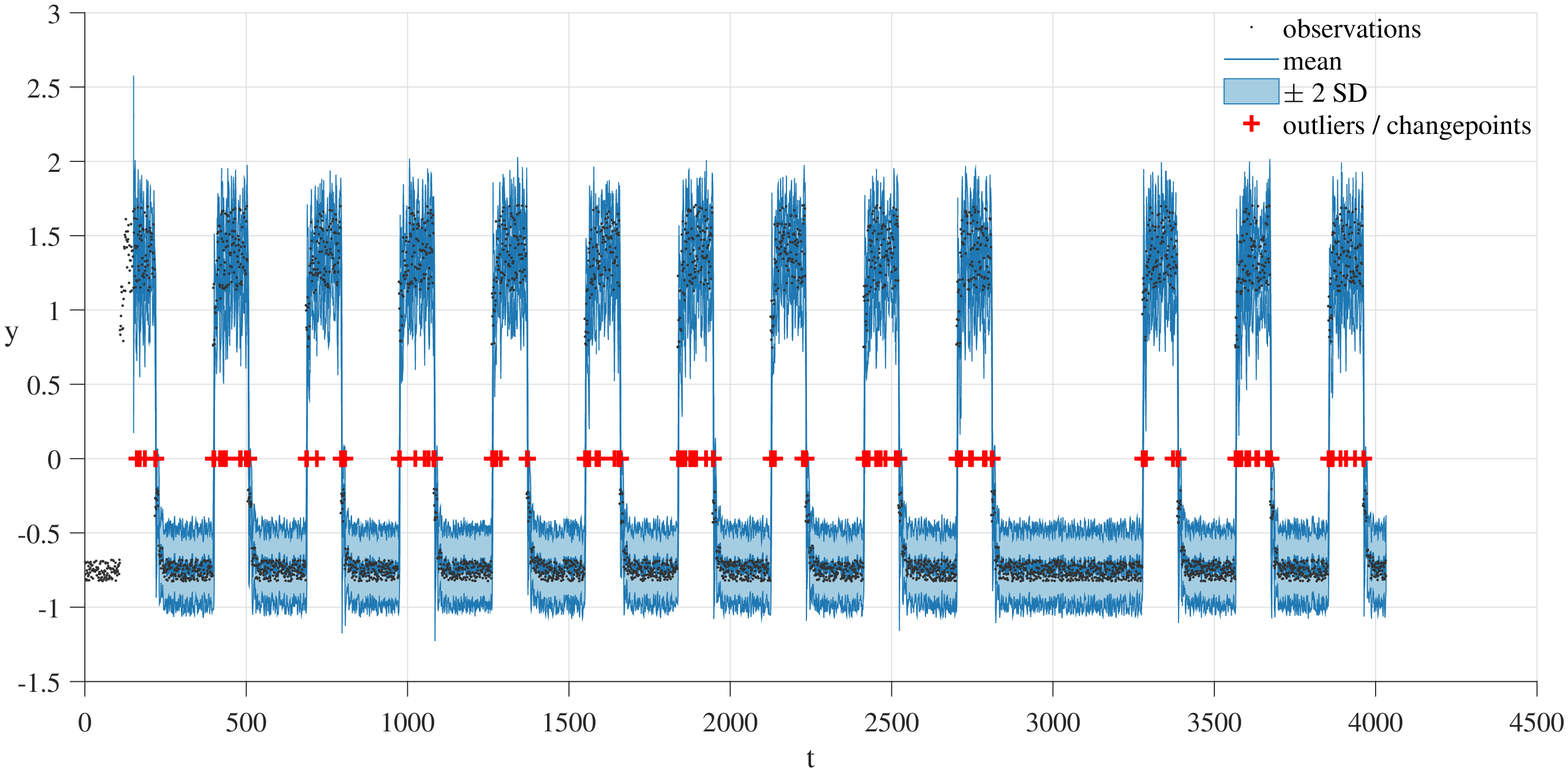}\\
\includegraphics[width=3.5in,height=1.5in]{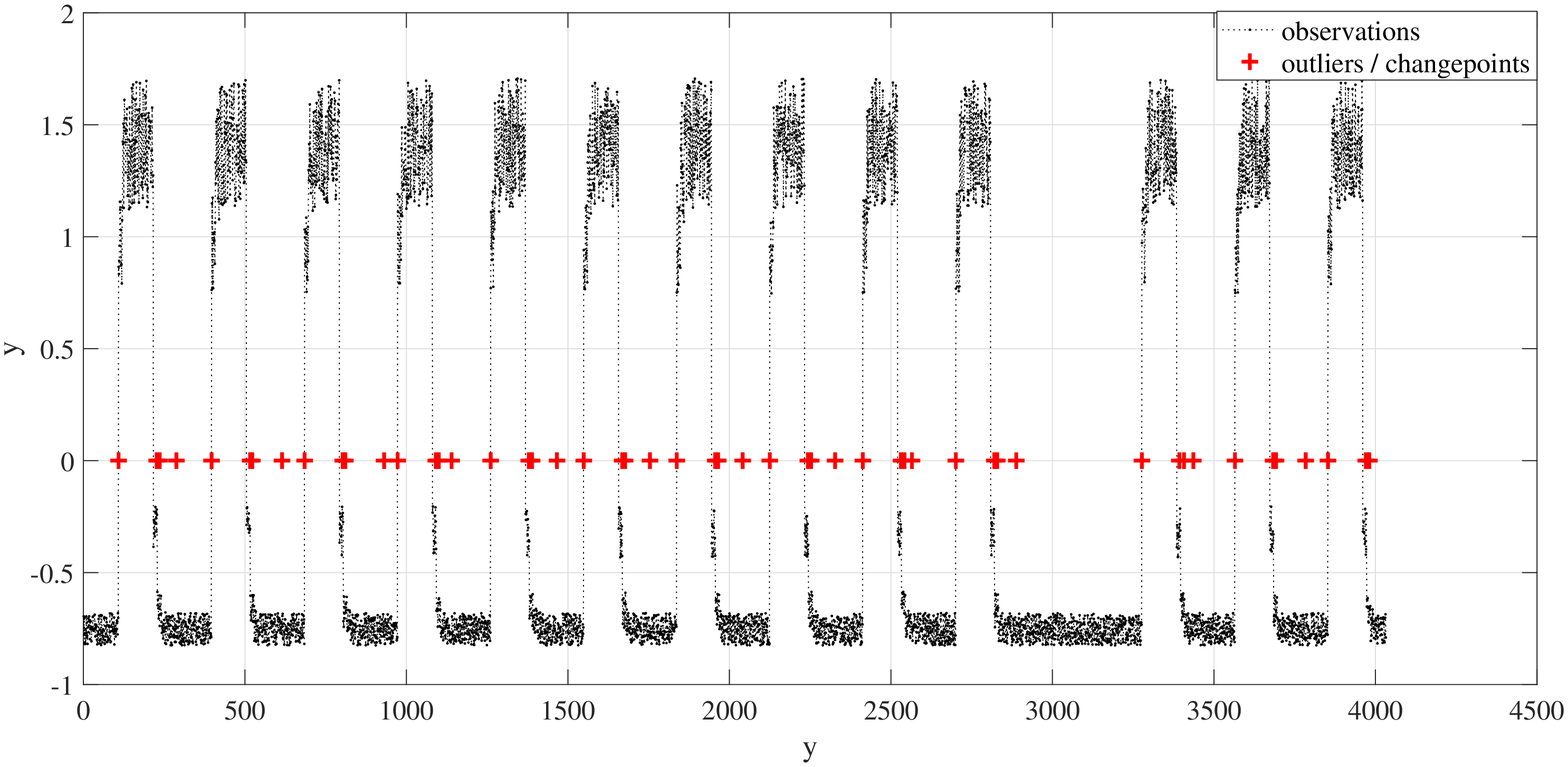}\\
\includegraphics[width=3.5in,height=1.5in]{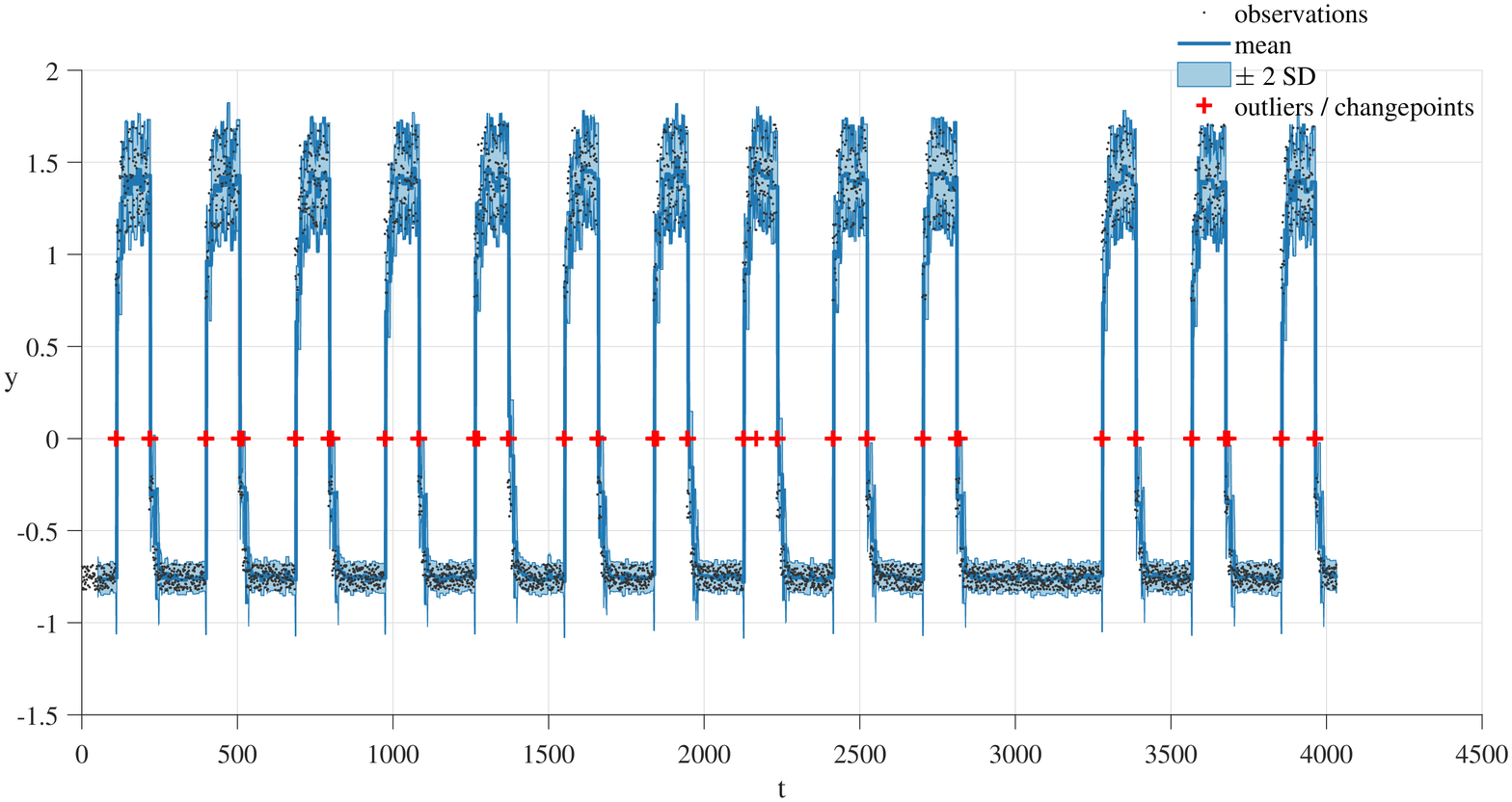}
\caption{Sequential real-time change point detection for a Numenta benchmark dataset \cite{lavin2015evaluating}. The upper, the middle and the bottom panel presents results corresponding to the fault bucket algorithm \cite{osborne2011machine}, the BOCPD algorithm \cite{turner2009adaptive}, and our proposed INTEL algorithm, respectively. The first 50 data points are used for hyper-parameter initialization.}\label{fig:art_daily_nojump}
\end{figure}

This dataset is included in the Numenta Anomaly Benchmark \cite{lavin2015evaluating}. It is characterized by a pattern of repeated amplitude changes, see Figure \ref{fig:art_daily_nojump}, hence is suitable for testing change points detection algorithms. We run fault bucket, BOCPD, and our INTEL algorithms, respectively, to process this dataset. The result is visually plotted in Figure \ref{fig:art_daily_nojump}. As is shown, all algorithms considered here have successfully detected the true change points, while both fault bucket and BOCPD give some false anomaly detections. In contrast, INTEL gives no false detection for this dataset. Besides, compared with the fault bucket algorithm, INTEL gives tighter $\pm2$SD bounds in its predictions.
\subsubsection{Fish killer data}
\begin{figure}
\centering
\includegraphics[width=3.5in,height=1.5in]{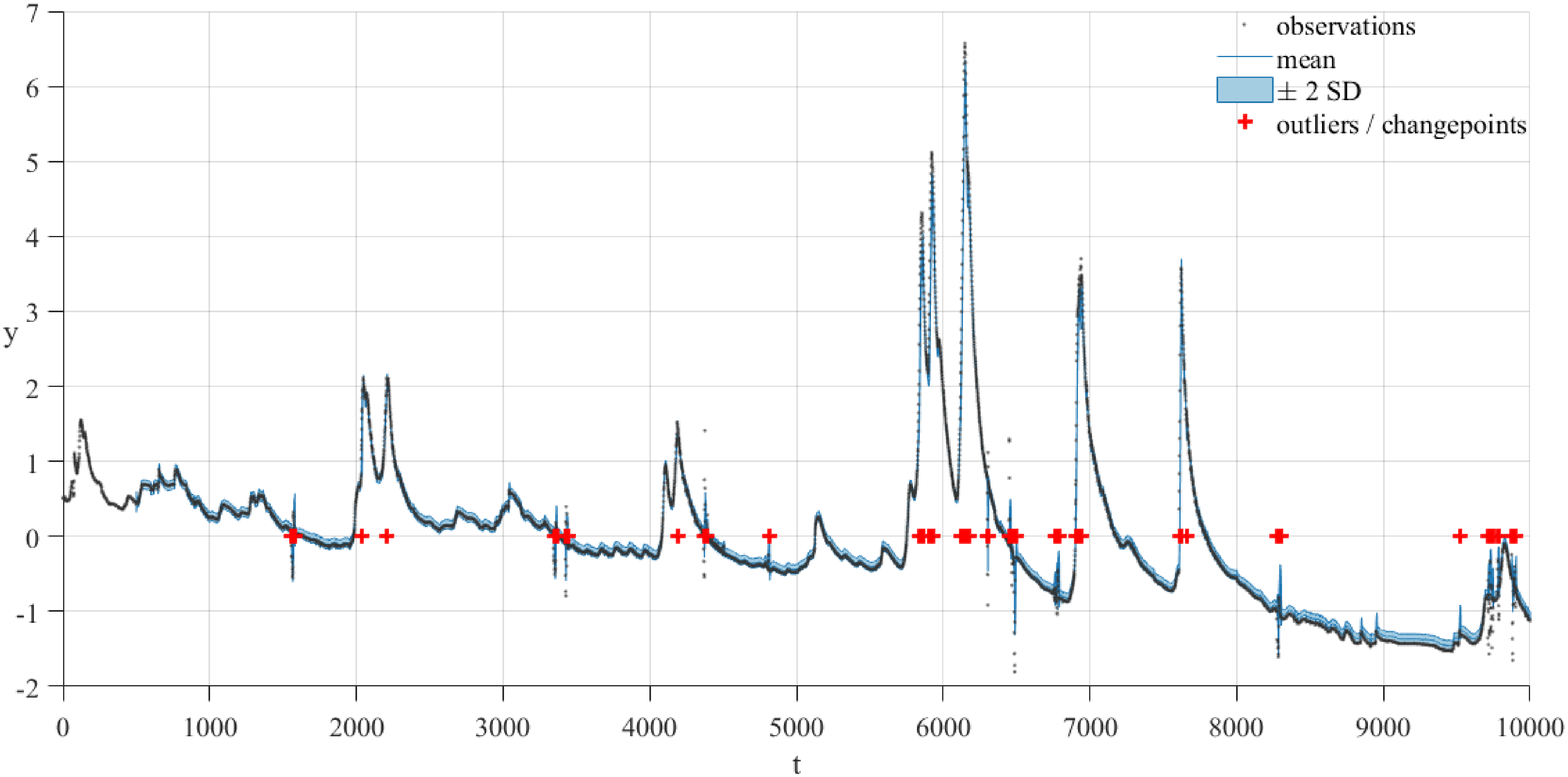}\\
\includegraphics[width=3.5in,height=1.5in]{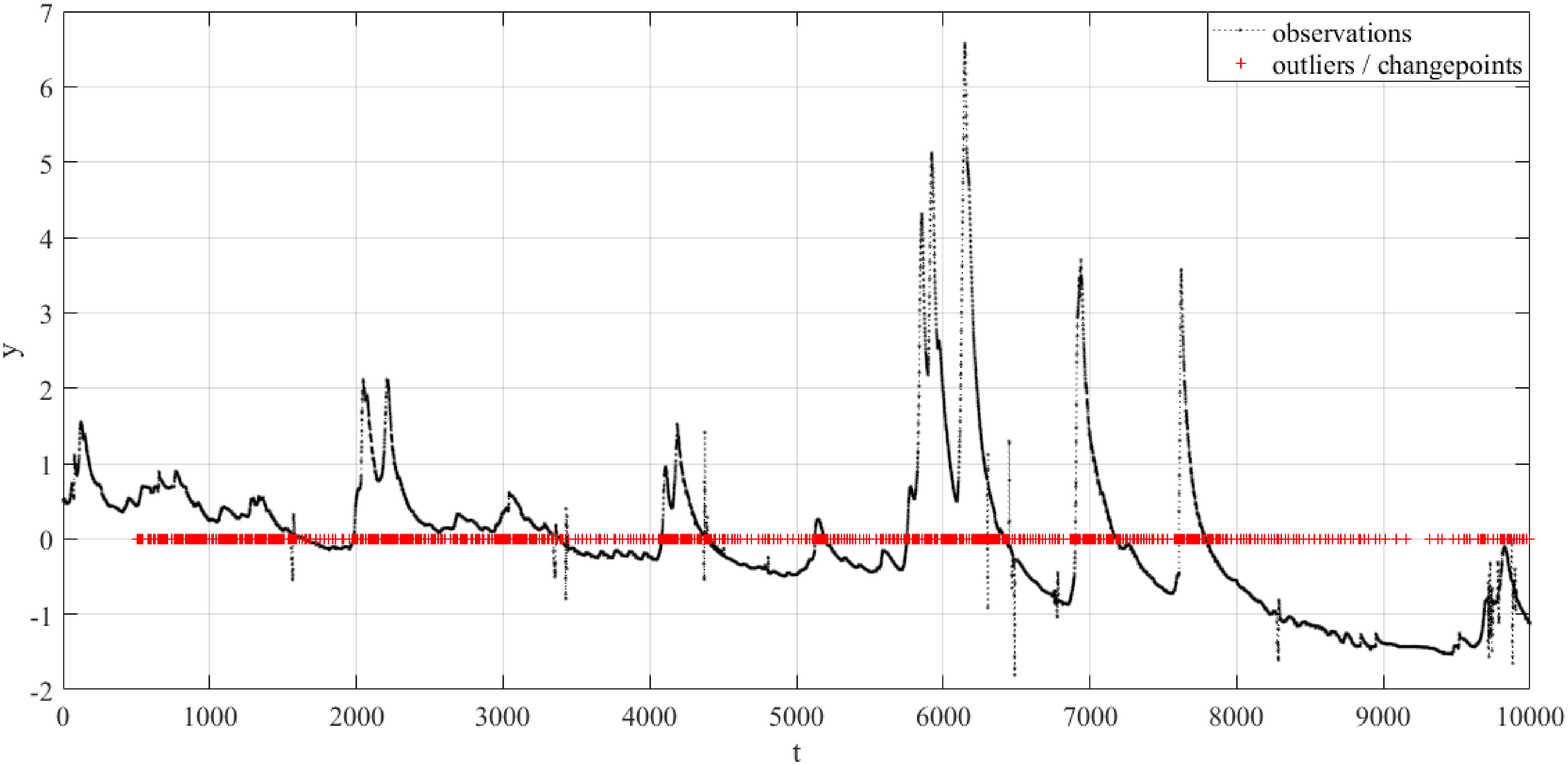}\\
\includegraphics[width=3.5in,height=1.5in]{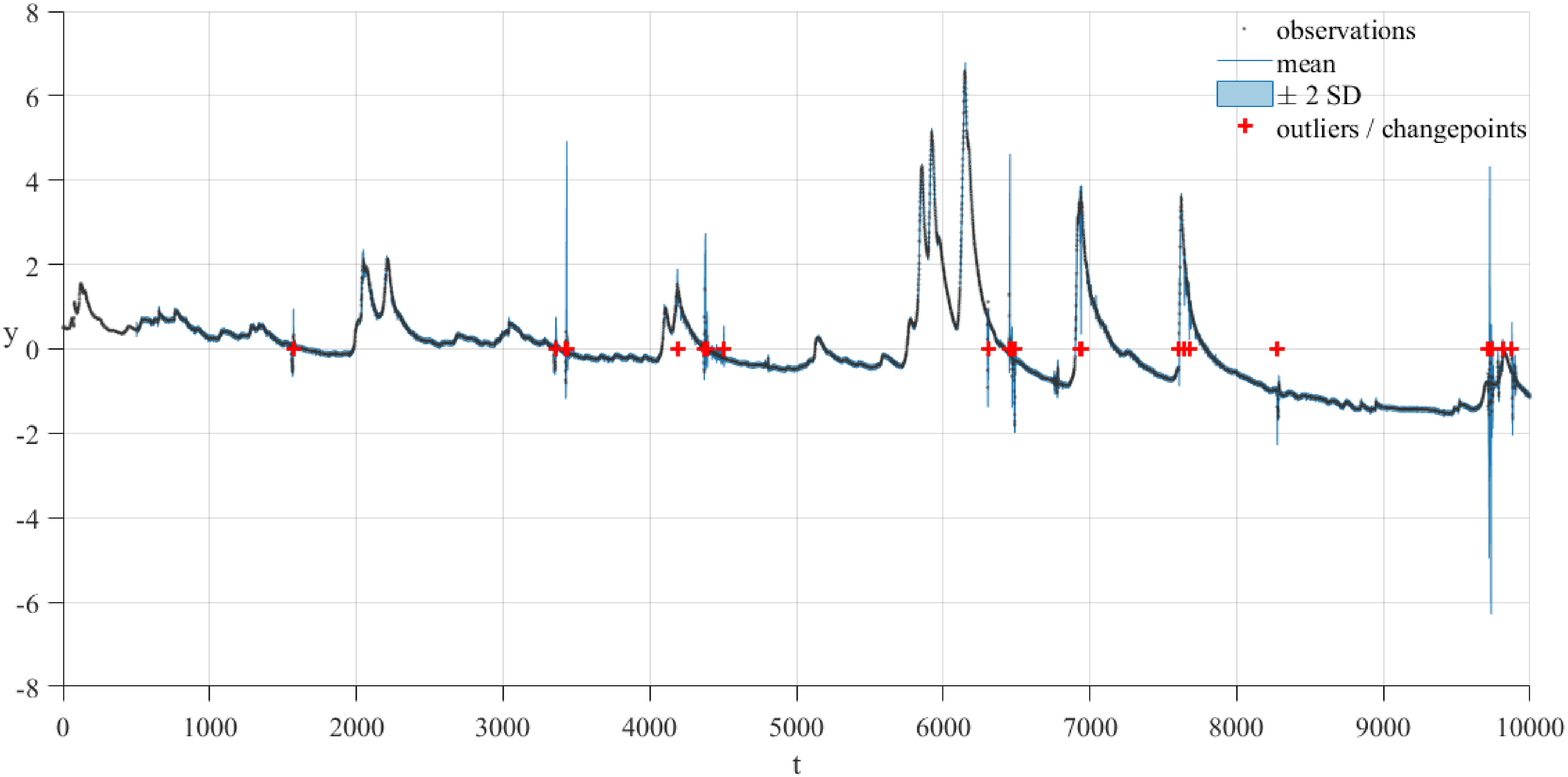}
\caption{Sequential real-time change point detection for the fish killer dataset. The upper, the middle and the bottom panels give results corresponding to the fault bucket algorithm \cite{osborne2011machine}, the BOCPD algorithm \cite{turner2009adaptive}, and our proposed INTEL algorithm, respectively. The first 500 data points are used for hyper-parameter initialization.}\label{fig:fishkiller}
\end{figure}

This dataset is a smooth time-series with some rapid changes near the fish kills. We selected the first 10,000 data points for use in comparing the fault bucket algorithm, BOCPD, and INTEL. The result is depicted in Figure \ref{fig:fishkiller}. It is shown that, for this dataset, fault bucket and INTEL give comparable one-step-ahead predictions. BOCPD performs unsatisfactorily since it reports many false anomaly detections.
\subsubsection{An industry portfolio data}
\begin{figure}
\centering
\includegraphics[width=3.5in,height=1.5in]{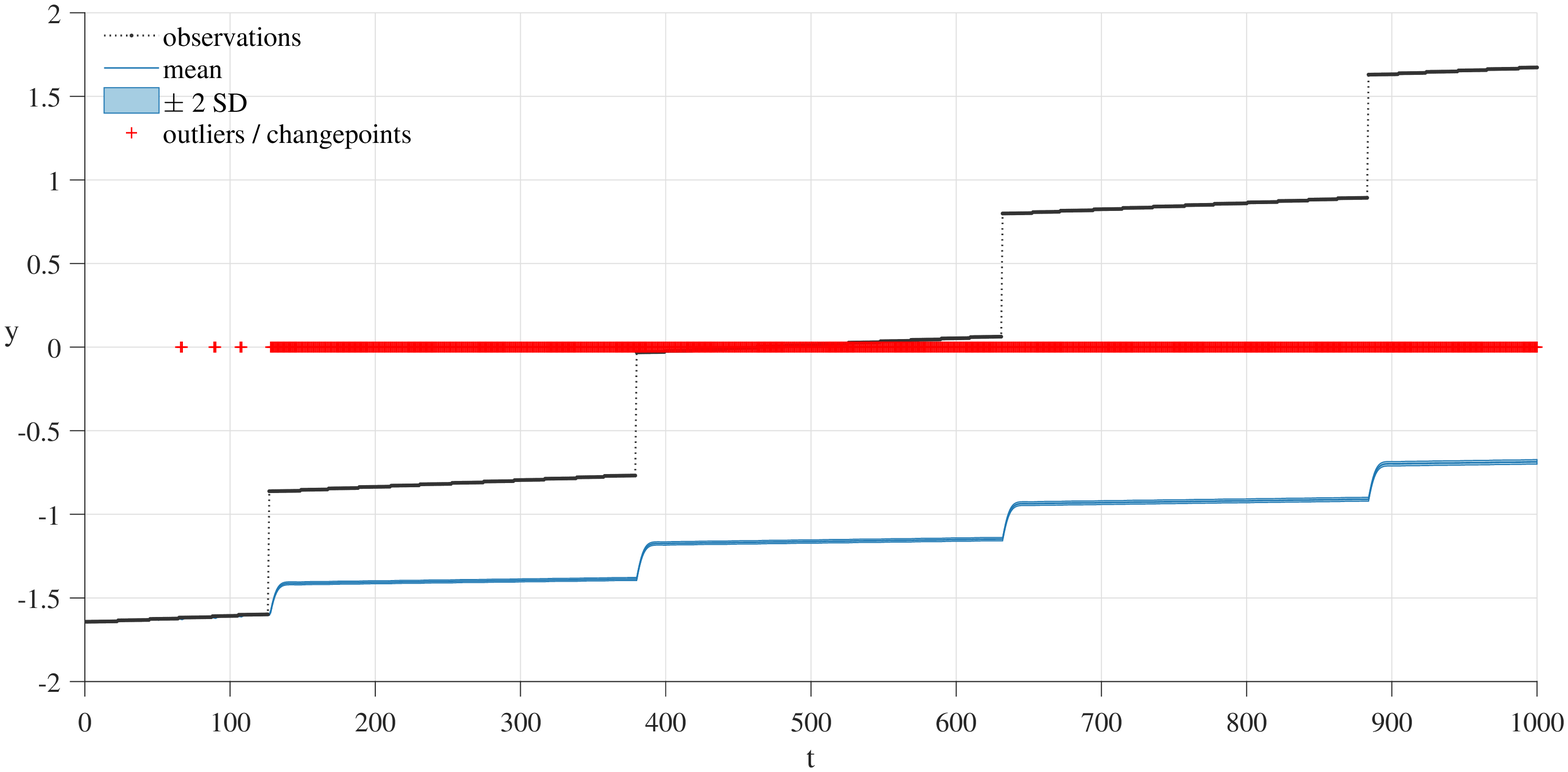}\\
\includegraphics[width=3.5in,height=1.5in]{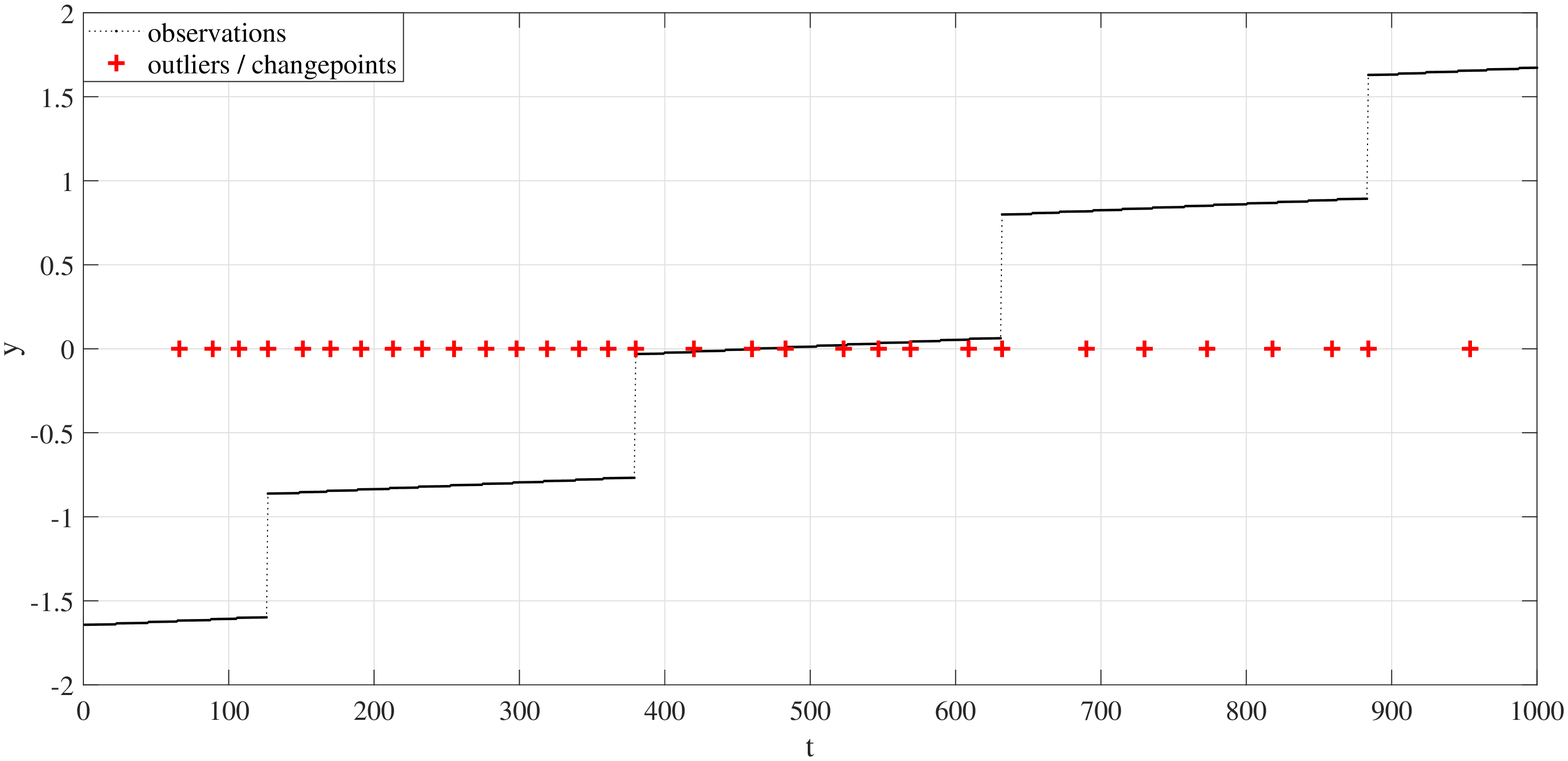}\\
\includegraphics[width=3.5in,height=1.5in]{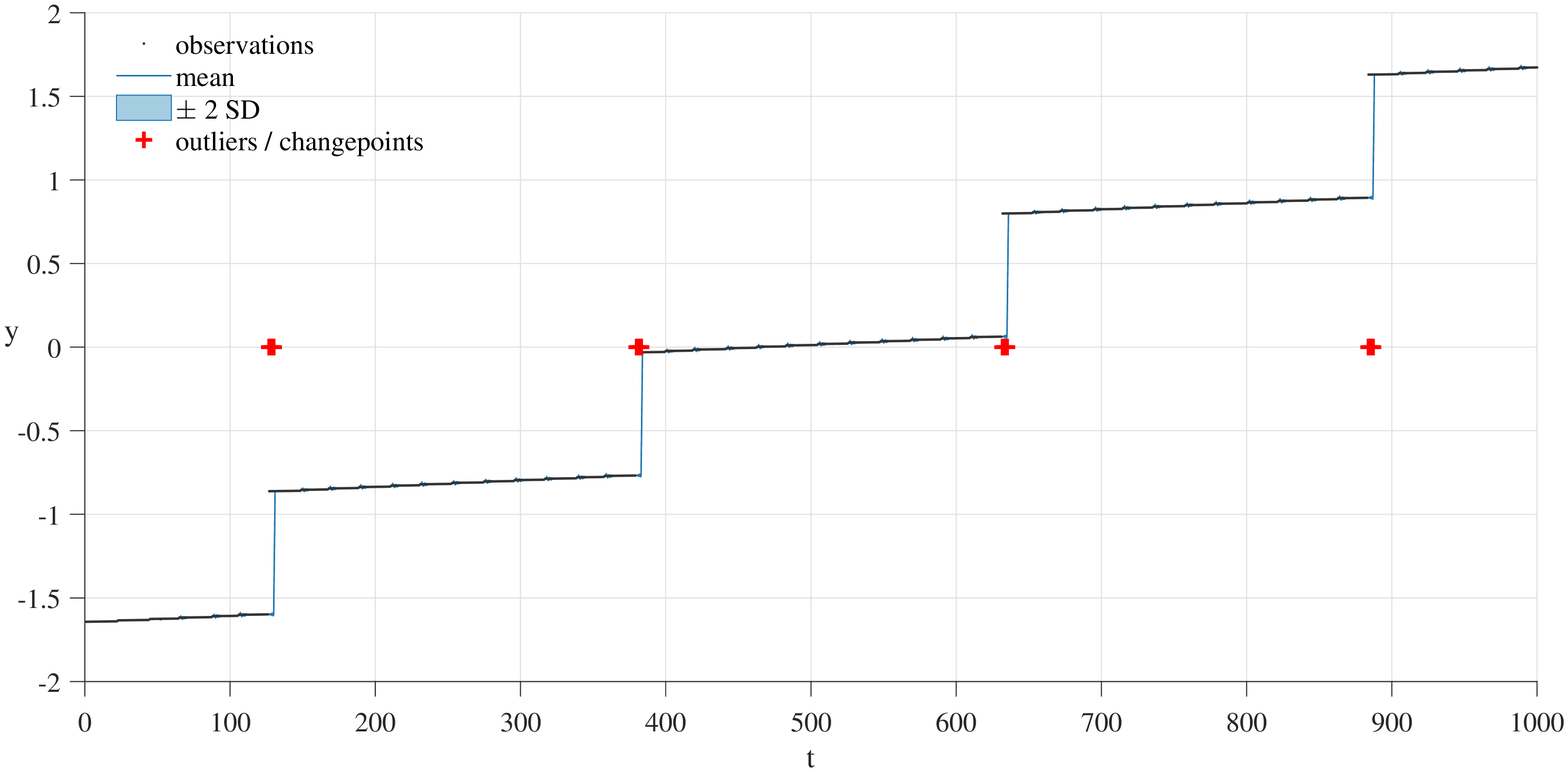}
\caption{Sequential real-time change point detection for an industry portfolio dataset. The upper, the middle and the bottom panels give results corresponding to the fault bucket algorithm \cite{osborne2011machine}, the BOCPD algorithm \cite{turner2009adaptive}, and our proposed INTEL algorithm, respectively. The first 50 data points are used for hyper-parameter initialization.}\label{fig:portfolio}
\end{figure}

We also considered the ``30 industry portfolios" dataset \cite{xuan2007modeling}. We selected a portion of the first time-series included in that dataset, which records daily returns of an industry-specific portfolio beginning at the year of 1963. The experimental result is plotted in Figure \ref{fig:portfolio}, from which one can see that our INTEL algorithm detects all change points accurately without any false detection. The fault bucket algorithm fails to yield accurate change point detections and observation predictions after the first regime shift. BOCPD successfully detects all change points, while it also declares many false detections.
\subsection{Experiments for prediction performance evaluation}\label{sec:pred_comp}
We tested the one-step-ahead prediction performance of our INTEL algorithm. Except for the fault bucket algorithm \cite{osborne2011machine}, we also included a simplified version of the INTEL algorithm, termed S-INTEL here. In S-INTEL, only the template model $\mathcal{M}_0$ is used, while the operations for anomaly detection, training set formation and hyper-parameter value adaptation maintain the same as that in INTEL. The BOCPD algorithm used in subsection \ref{sec:cpd_detect} is not involved here since it is only capable of detecting change points but incapable of making real-time predictions. Except those used in subsection \ref{sec:cpd_detect}, additional time-series datasets are considered here, including:
\begin{itemize}
\item the Nile dataset, which has been widely used in the time-series literature;
\item the Intel lab data, which was collected from 54 sensors deployed in the Intel Berkeley Research lab between Feb. 28th and April 5th, 2004. We only used a small while representative fragment of this dataset.
\item the NYC taxi data, which records the number of NYC taxi passengers. Each observation in this dataset denotes the total number of taxi passengers during 30 minutes. Five regime shifts happen during the NYC marathon, Thanksgiving, Christmas, New Years day, and a snow storm, respectively.
\item the temperature (temp.) sensor data of an internal component of a large, industrial machine. This dataset has at least two outlier observations. One originates from a planned shutdown of the machine, and the other one is a catastrophic failure of the machine.
\item A real time traffic data from the twin cities metro area in Minnesota of the U.S.. Included metrics include occupancy, speed, and travel time from specific sensors, while we only present the result associated with the metric speed, due to the limitation in space.
\end{itemize}

The performance metrics in use are the negative log likelihood (NLL), the mean absolute error (MAE), and the mean square error (MSE). For every metric, the smaller is its value, the better the prediction performance it stands for. We list the one-step-ahead prediction result measured with these metrics in Tables \ref{Table:NLL}-\ref{Table:MSE}.

\begin{figure}
\centering
\includegraphics[width=3.5in,height=1.5in]{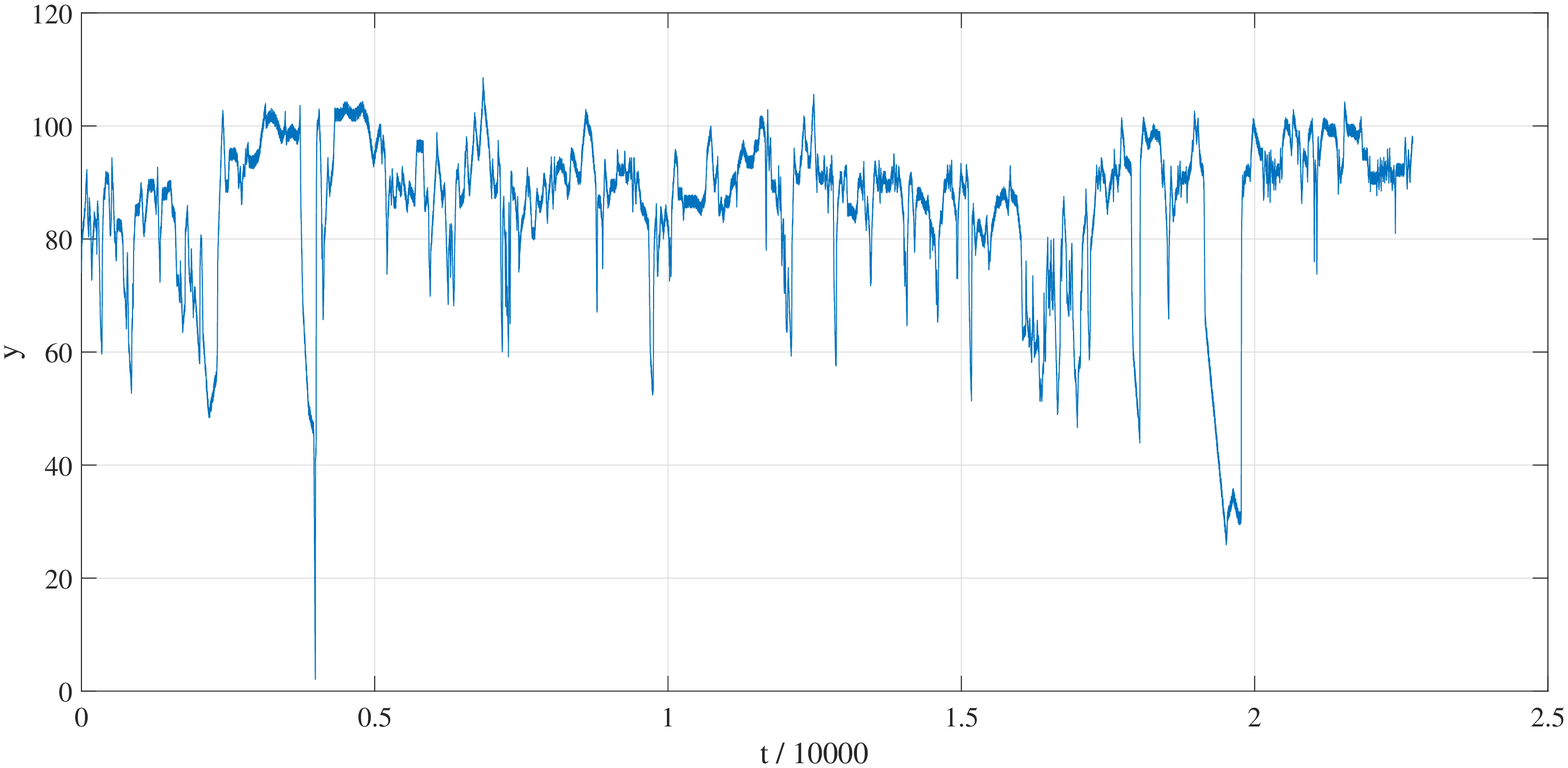}\\
\includegraphics[width=3.5in,height=1.5in]{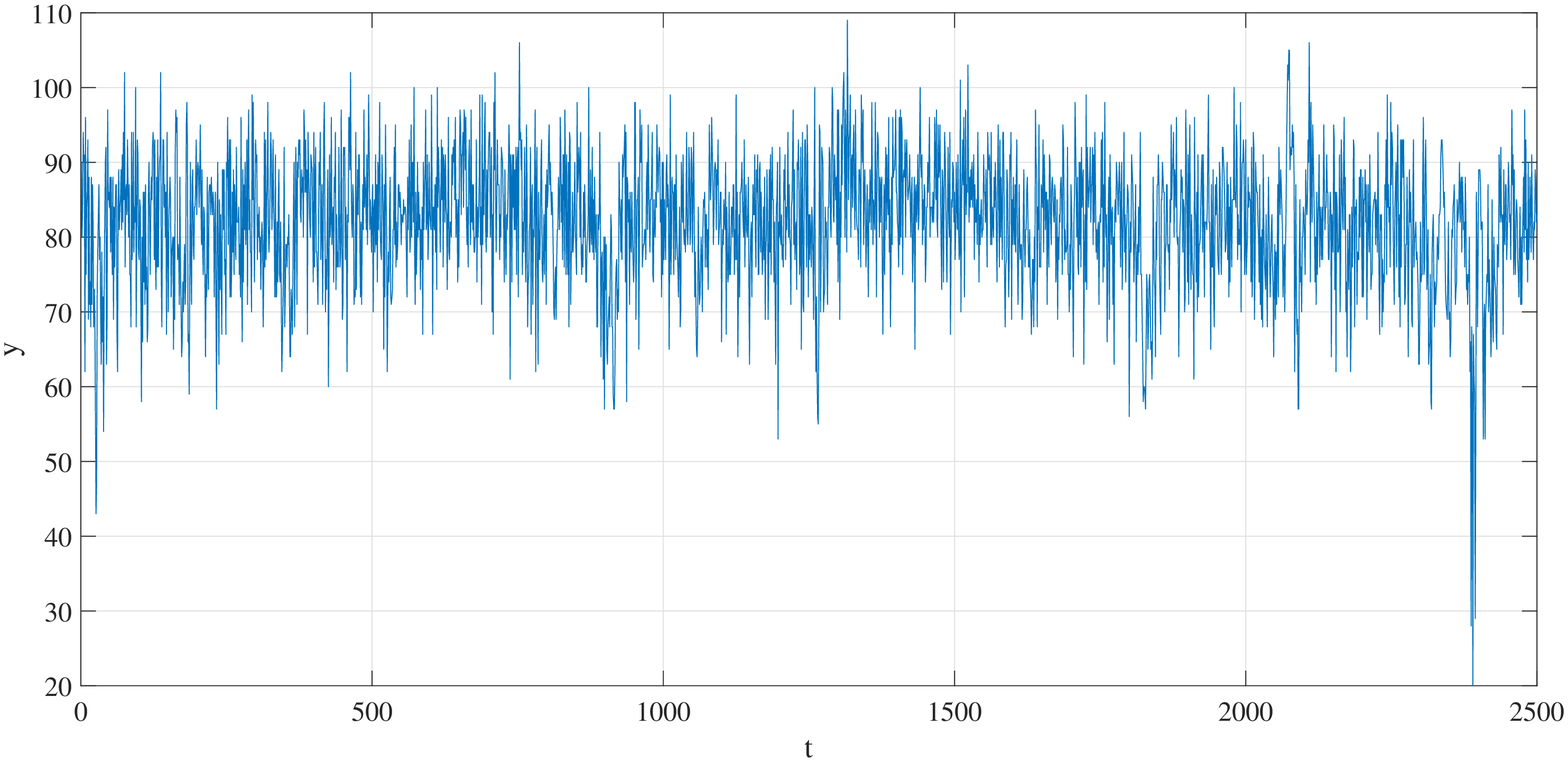}
\caption{The temp. sensor dataset (the top panel) and the traffic dataset (the bottom panel) used in Tables \ref{Table:NLL}-\ref{Table:MSE}}\label{fig:temp_traffic_data}
\end{figure}

As is shown in Tables \ref{Table:NLL}-\ref{Table:MSE}, for the first 8 of these 11 datasets, INTEL outperforms the fault bucket algorithm \cite{osborne2011machine} in terms of all metrics considered. For the last dataset, INTEL performs slightly better than the fault bucket algorithm in terms of MAE and MSE, while the fault bucket algorithm beats INTEL slightly in terms of NLL. It is only for the NYC taxi dataset and the temp. sensor dataset that the fault bucket algorithm gives significantly better prediction than INTEL. We plot these two datasets in Figure \ref{fig:temp_traffic_data}. As is shown, there is no clear regime shift in them. It indicates that the advantage of INTEL over the fault bucket algorithm mainly comes from its capability to handle change points.

By comparing S-INTEL and INTEL according to results as shown in Tables \ref{Table:NLL}-\ref{Table:MSE}, one can see that INTEL outperforms S-INTEL markedly in most cases. S-INTEL only provides slightly better prediction than INTEL in terms of MAE for the first two datasets, and in terms of MSE for the first dataset. The above result demonstrates the advantage of using multiple models compared with using only one model.
\begin{table}\centering\small
\caption{NLL based prediction performance comparison}
\begin{tabular}{c|c|c|c}
\hline %
  & Fault bucket & S-INTEL & INTEL\\\hline
CPU usage & 3.5181 & 0.7785 & \textbf{0.0972}\\\hline
well-log & 46.3338 & 0.0950 & \textbf{0.0947} \\\hline
ECG & 24.6629 & 42.2999 & \textbf{-0.1459} \\\hline
Numenta & -1.0409 & -0.4663 & \textbf{-1.4887} \\\hline
fish killer & -1.0336 & 6.2875 & \textbf{-1.6867} \\\hline
portfolio & 60,200 & 8.8275 & \textbf{-3.7451} \\\hline
Nile data & 129.4349 & 22.8654 & \textbf{2.2453}  \\\hline
Intel lab & -0.8593 & -0.6276 & \textbf{-1.2252} \\\hline
NYC taxi & \textbf{-0.4434} & 5.6331 & 0.0129 \\\hline
temp. sensor & \textbf{-0.7499} & 41.3108 & -0.2588 \\\hline
traffic & \textbf{1.3471} & 1.3525 & 1.3749 \\\hline
\end{tabular}
\label{Table:NLL}
\end{table}
\begin{table}\centering\small
\caption{MAE based prediction performance comparison}
\begin{tabular}{c|c|c|c}
\hline %
  & Fault bucket & S-INTEL & INTEL\\\hline
CPU usage & 1.9540 & \textbf{0.1859} & 0.1867\\\hline
well-log & 2.4985 & \textbf{0.2078} & 0.2129 \\\hline
ECG & 0.4650 & 0.3786 & \textbf{0.1387} \\\hline
Numenta & \textbf{0.0375} & 0.0527 & 0.0525 \\\hline
fish killer & 0.0496 & 0.0478 & \textbf{0.0220} \\\hline
portfolio & 2.3378 & 0.0034 & \textbf{0.0012} \\\hline
Nile & 1.8076 & 0.6611 & \textbf{0.6111} \\\hline
Intel lab & 0.0833 & 0.0628 & \textbf{0.0485} \\\hline
NYC taxi & \textbf{0.1102} & 0.3761 & 0.1943 \\\hline
temp. sensor & \textbf{0.0894} & 0.4028 & 0.1481 \\\hline
traffic & 0.6716 & 0.6725 & \textbf{0.6500} \\\hline
\end{tabular}
\label{Table:MAE}
\end{table}
\begin{table}\centering\small
\caption{MSE based prediction performance comparison}
\begin{tabular}{c|c|c|c}
\hline %
  & Fault bucket & S-INTEL & INTEL\\\hline
CPU usage & 3.8726 & \textbf{0.0562} & 0.0564\\\hline
well-log & 6.3093 & 0.0707 & \textbf{0.0706} \\\hline
ECG & 1.5398 & 1.0580 & \textbf{0.0404} \\\hline
Numenta & \textbf{0.0044} & 0.0057 & 0.0057 \\\hline
fish killer & \textbf{0.0129} & 0.0208 & 0.0182 \\\hline
portfolio & 5.4651 & $4.4059\times 10^{-5}$ & \textbf{$4.8654\times 10^{-6}$} \\\hline
Nile & 3.5819 & 0.6604 & \textbf{0.5535} \\\hline
Intel lab & 0.0073 & 0.0059 & \textbf{0.0038} \\\hline
NYC taxi & \textbf{0.0184} & 0.2045 & 0.0635 \\\hline
temp. sensor & \textbf{0.0101} & 0.3342 & 0.0313 \\\hline
traffic & 0.7103 & 0.7263 & \textbf{0.6903} \\\hline
\end{tabular}
\label{Table:MSE}
\end{table}
\subsection{Robustness test}\label{sec:robust_test}
\begin{figure}
\centering
\includegraphics[width=3.5in,height=1.5in]{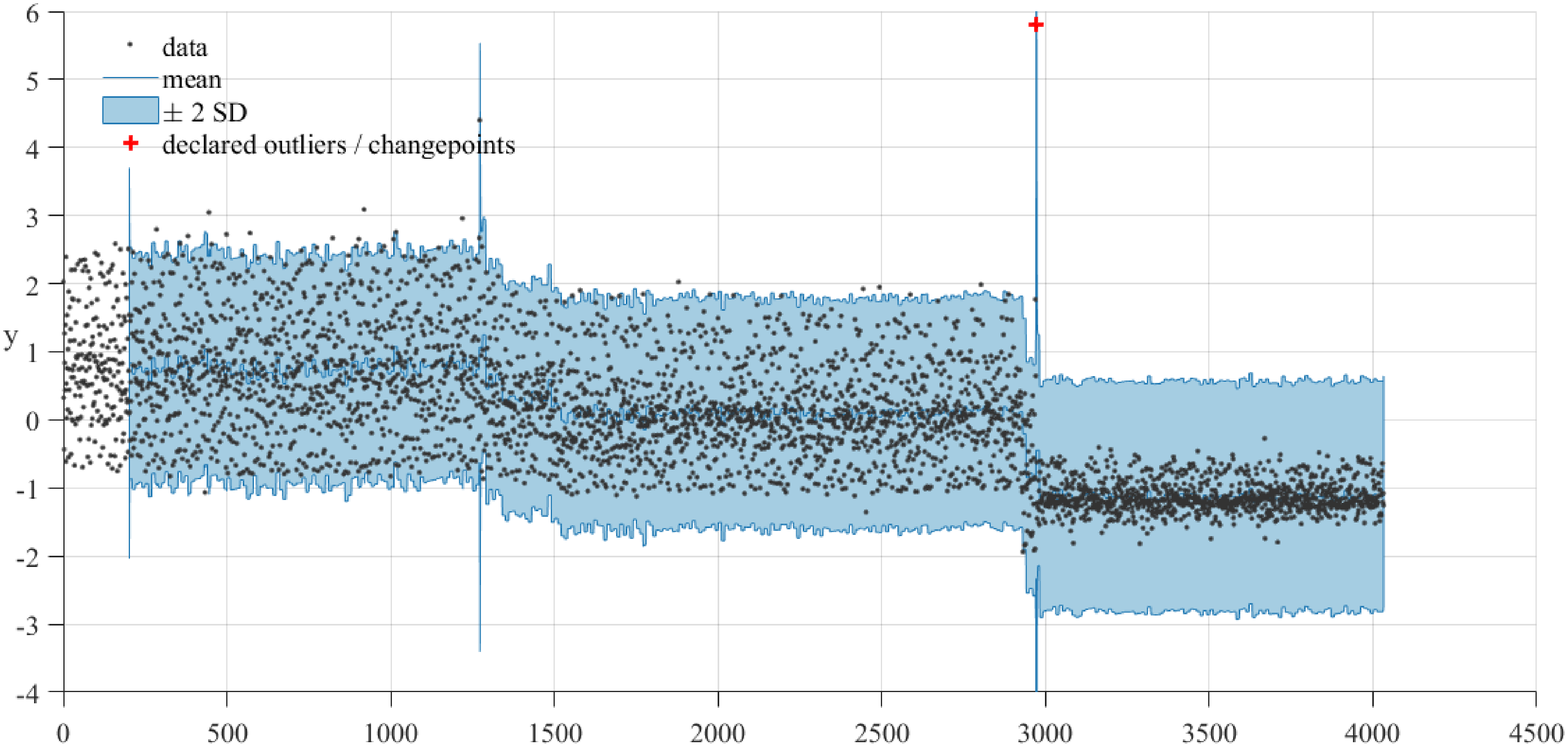}\\
\includegraphics[width=3.5in,height=1.5in]{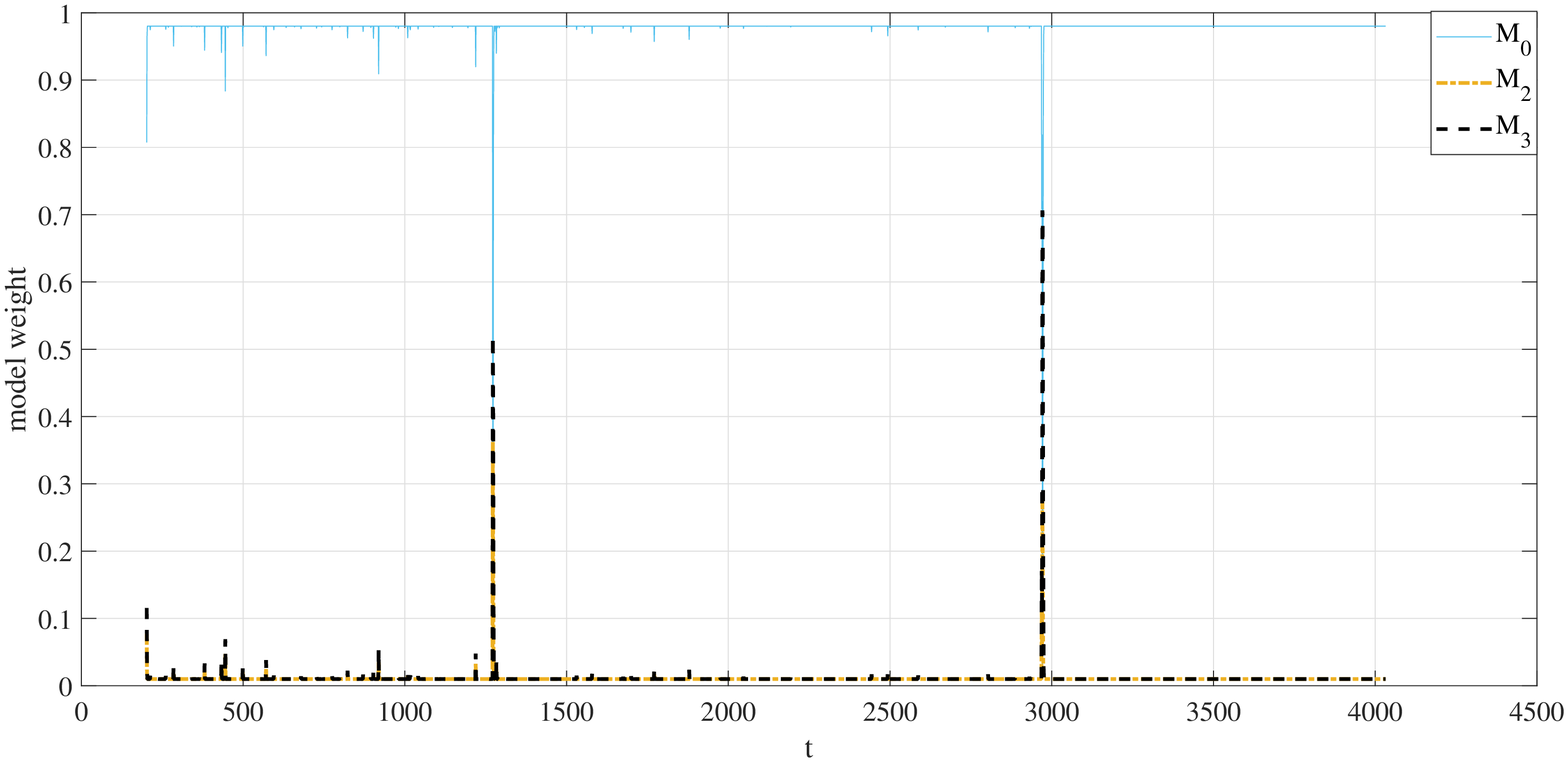}\\
\includegraphics[width=3.5in,height=1.5in]{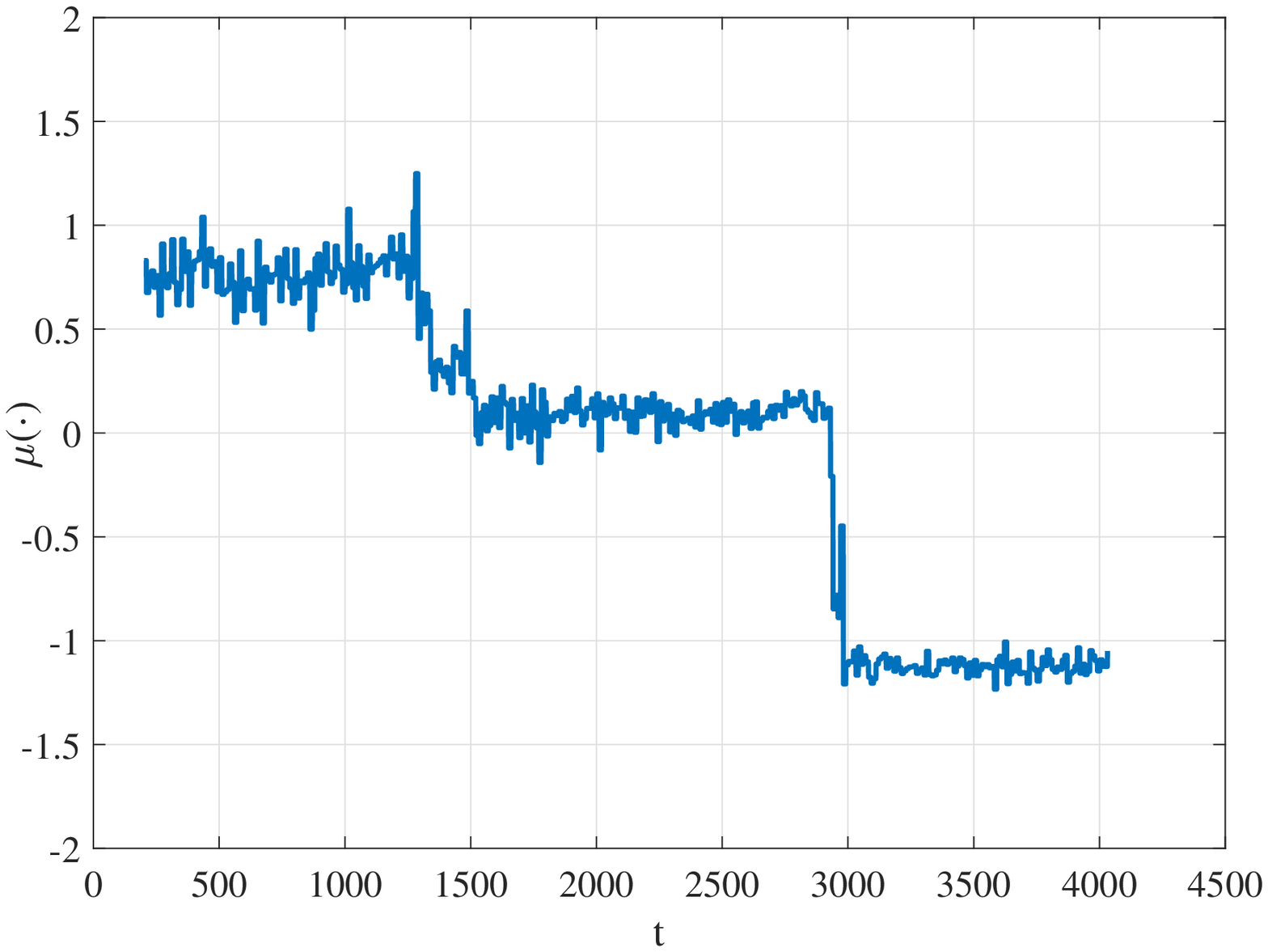}
\caption{Experimental result of the INTEL algorithm when adopting an inaccurate prior knowledge for initializing $\mathcal{M}_i, i>0$. The experimental setting is the same as that used for plotting Figure \ref{fig:ec2_cpu_data_pred_init2}, except that $\mathcal{M}_1$ is now removed from the model set. The same as in Figure \ref{fig:ec2_cpu_data_pred_init2}, the middle panel shows that the INTEL algorithm assigns for $\mathcal{M}_2$ and $\mathcal{M}_3$ tiny weight values almost all the time, except at $t=1272, 2971$, where a regime shift happens.}\label{fig:ec2_cpu_data_pred_init3}
\end{figure}

We tested the robustness of our INTEL algorithm in three cases as below:
\begin{enumerate}
\item The prior knowledge used for initializing $\mathcal{M}_i, i>0$ is inaccurate;
\item The historical dataset used for initializing $\mathcal{M}_0$ is not clean, namely, there is at least one outlier or change point included in it.
\item False detections of anomalies exist during the sequential prediction process.
\end{enumerate}
For case 1 listed above, we modified the initialization setting used in subsection \ref{sec:test_init} for processing the CPU usage dataset. Specifically, $\mathcal{M}_1$ is removed from the model set, while $\mathcal{M}_0$, $\mathcal{M}_2$ and $\mathcal{M}_3$ remain. Recall that in $\mathcal{M}_2$ and $\mathcal{M}_3$, we have $\sigma_{f,2}=15\sigma_{f,0}$ and $\sigma_{f,3}=10\sigma_{f,0}$, respectively. We now adopt an inaccurate prior knowledge that the observation amplitude will increase during some period but never decrease, while the fact is that it will decrease significantly after $t=2,971$. The performance of INTEL under this setting is plotted in Figure \ref{fig:ec2_cpu_data_pred_init3}. We see that the INTEL algorithm fails to capture one important aspect of the temporal structure, namely a significantly lowered amplitude, in the data after the regime shift at $t=2,791$. However, it still gives accurate mean predictions for observations after $t=2,791$.
\begin{table}[!htb]\centering\small
\caption{Prediction performance of the proposed INTEL algorithm for case 2 in subsection \ref{sec:robust_test}}
\begin{tabular}{c|c|c}
\hline %
NLL  & MAE & MSE \\\hline
0.5545 & 0.2126 & 0.0708\\\hline
\end{tabular}
\label{Table:robust_test}
\end{table}
\begin{table}[!htb]
\centering\small
\caption{Prediction performance of the proposed INTEL algorithm for case 3 in subsection \ref{sec:robust_test}}
\begin{tabular}{c|c|c}
\hline %
NLL  & MAE & MSE \\\hline
0.5008 & 0.2150 & 0.0756\\\hline
\end{tabular}
\label{Table:robust_test_case3}
\end{table}
\begin{figure}
\centering
\includegraphics[width=3.5in,height=1.5in]{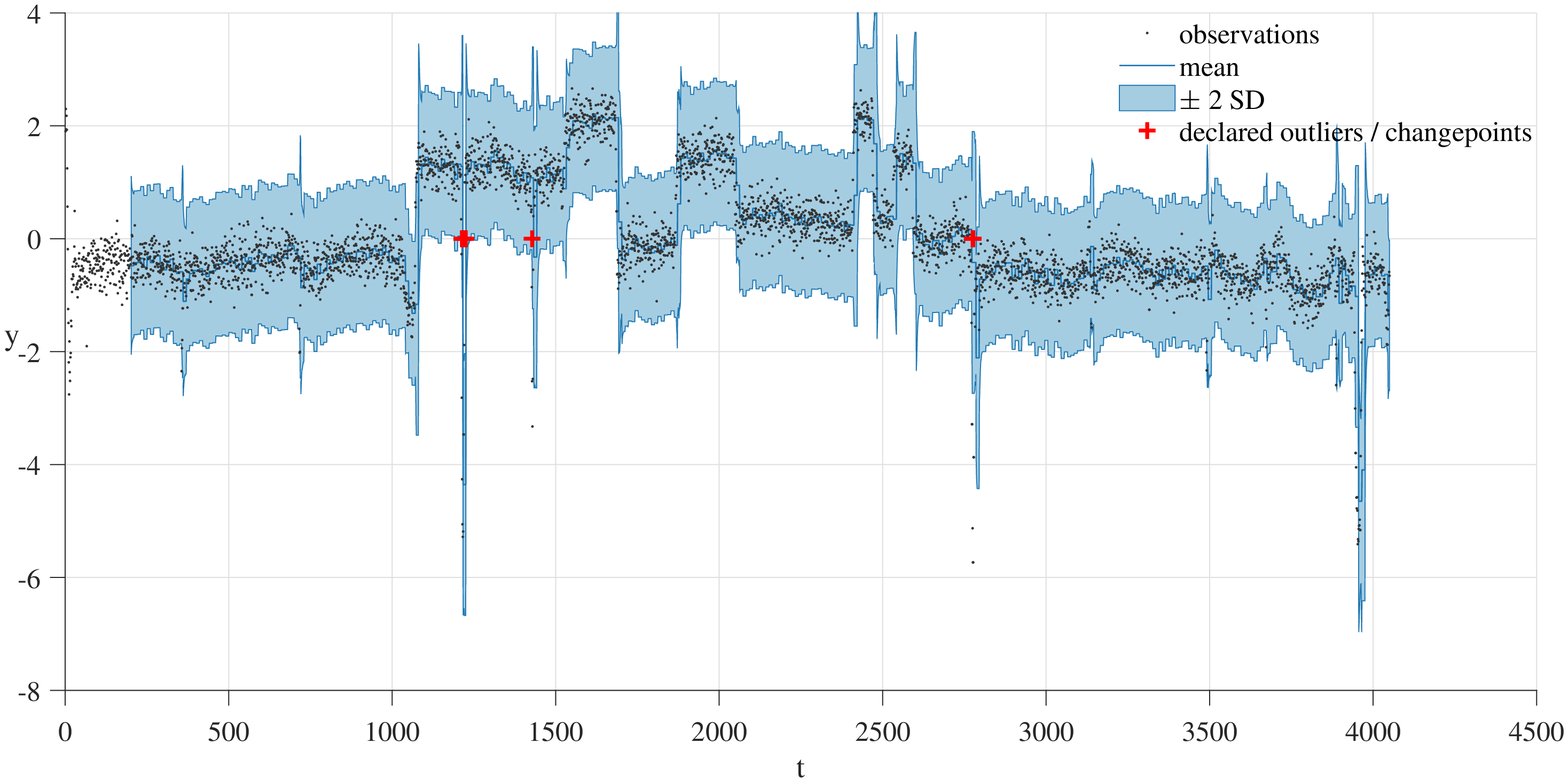}
\caption{Sequential prediction result of the proposed INTEL algorithm for the well-log dataset. The first 200 data points, in which anomaly observations are present, are used for hyper-parameter initialization. The other experimental settings are the same as that used for plotting Figure \ref{fig:well}.}\label{fig:well2}
\end{figure}

For case 2, we re-studied the well-log dataset. In the result plotted in Figure \ref{fig:well}, data points between $t=100$ and $t=300$ are used for hyper-parameter initialization for $\mathcal{M}_0$, since there is no anomaly observation within them. We now use the first 200 data points for hyper-parameter initialization for $\mathcal{M}_0$. All the other experimental settings are kept the same as that used for plotting Figure \ref{fig:well}. Now anomalies exist in the training dataset (there are at least three anomalies in the first 50 data points as reported by \cite{turner2012gaussian}). Now the performance of the proposed INTEL algorithm is plotted in Figure \ref{fig:well2}. Comparing Figure \ref{fig:well2} with the bottom panel of Figure \ref{fig:well}, we see that INTEL fails to detect some change points now, and gives a broader $\pm2$SD bounds, while it still provides accurate mean predictions. The result presented in Table \ref{Table:robust_test} reconfirms the above observation. Comparing the result listed in Table \ref{Table:robust_test} with that shown in Tables \ref{Table:NLL}-\ref{Table:MSE}, we see that the performance of INTEL is almost unchanged in terms of MAE and MSE. Its performance is degraded based only on the metric NLL. This is because the metrics MAE and MSE only describe the accuracy of the mean prediction, while NLL covers information on the uncertainty measure.
\begin{figure}
\centering
\includegraphics[width=3.5in,height=1.5in]{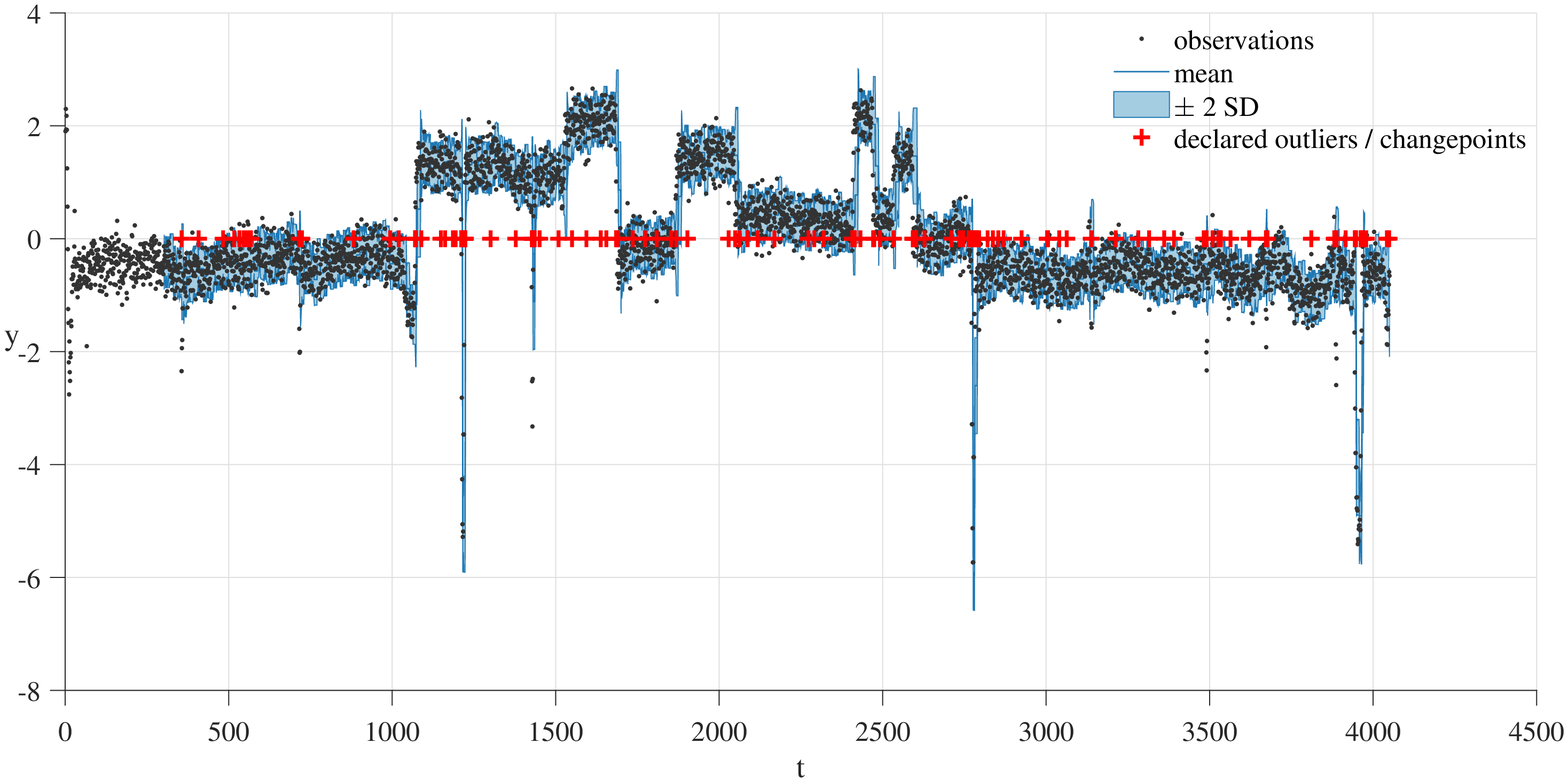}
\caption{Sequential prediction result of the proposed INTEL algorithm for the well-log dataset, in case of false anomaly detections being present.}\label{fig:well3}
\end{figure}

Finally, for case 3, we tested the performance of INTEL when false anomaly detections are present for the well-log data. As  is shown in Figure \ref{fig:well3}, even in case of false anomaly detections being present, the INTEL algorithm can still give accurate predictions for most of the observations. A quantitative evaluation of the prediction performance associated with Figure \ref{fig:well3} is shown in Table \ref{Table:robust_test_case3}. Comparing Table \ref{Table:robust_test_case3} with that shown in Tables \ref{Table:NLL}-\ref{Table:MSE}, again, we see that the performance of INTEL is degraded based only on the metric NLL.
\section{Concluding remarks and future works}\label{sec:con}
In this paper, we addressed the problem of SOP by unleashing the flexibility and interpretability of the GPTS model together with harnessing prior knowledge. Specifically, we proposed a novel algorithm design termed INTEL and demonstrated its performance using extensive real dataset experiments. Experimental results show that the INTEL algorithm is a highly efficient solution to the problem of SOP in the presence of outliers and change points. As an online prediction algorithm, INTEL is also demonstrated to be a qualified online anomaly detection method. The biggest feature of INTEL is that it can instantly capture the pattern of the new regime, without the need to do model training, upon a change point is declared. Further, the INTEL algorithm allows closed-form inference and prediction. All operations to implement this algorithm are deterministic and analytically tractable.

We did robustness tests to the INTEL algorithm, investigating its performance under three undesirable cases, namely, when the prior knowledge it adopts is inaccurate, when the historical data used for template model hyper-parameter initialization is not clean and when false anomaly detections exist during the sequential prediction process. An interesting finding is that, under these cases, although the INTEL's prediction performance is degraded in terms of the metric NLL, its prediction performance in terms of MAE and MSE maintains. That says our INTEL algorithm can still provide accurate point predictions in our test cases.

Currently, the INTEL algorithm can only do one-step-ahead prediction, while, in principle, it can be extended naturally to do multiple-step-ahead prediction, which deserves future investigation. It is also important to extend the INTEL algorithm to handle multi-variate time-series data. In the current version of the INTEL algorithm, each candidate GPTS model uses a Matern 5/2 kernel function. It is possible to let these candidate models employ different types of kernel functions and then check its performance.




\bibliographystyle{IEEEbib}
\bibliography{mybibliography}
\end{document}